\documentclass[journal]{IEEEtran}
\usepackage{amsmath,amsthm,amssymb,amsfonts}
\usepackage{enumitem}
\usepackage{dsfont}
\usepackage{soul}
\usepackage{mathtools}
\usepackage[utf8]{inputenc}
\usepackage{multirow}
\usepackage[colorlinks]{hyperref}
\usepackage{cleveref}
\usepackage{bbm}
\usepackage{algorithm,algorithmic}
\usepackage{multirow}
\usepackage[normalem]{ulem}
\usepackage{makecell}
\newcommand{\beq}{ \begin{equation} }
\newcommand{\eeq}{ \end{equation} }

\usepackage{xr}
\usepackage{xr-hyper}
\usepackage[colorlinks]{hyperref}
\usepackage{cleveref}

\newtheorem{lem}{Lemma}[section]

\newtheorem{defi}{Definition}[section]
\newtheorem{thm}{Theorem}[section]
\newtheorem{cor}{Corollary}[section]

\newtheorem{rmk}{Remark}[section]

 \theoremstyle{remark}

\begin{document}
 
\title{Robust Hypergraph Clustering via\\ Convex Relaxation of Truncated MLE}
\author{Jeonghwan Lee, Daesung Kim and Hye Won Chung
\thanks{Jeonghwan Lee (sa8seung@kaist.ac.kr) is with the Department of Mathematical Sciences at KAIST; Daesung Kim (jklprotoss@kaist.ac.kr) and Hye Won Chung (hwchung@kaist.ac.kr) are with the School of Electrical Engineering at KAIST. This work was supported by National Research Foundation of Korea under grant number 2017R1E1A1A01076340, the Ministry of Science and ICT, Korea, under the ITRC support program (IITP-2020-2018-0-01402), and  by the Institute of Information \& Communications Technology Planning \& Evaluation (IITP) grant funded by the Korea government MSIT
(No. 2020-0-00626). Correspondence to: Hye Won Chung (hwchung@kaist.ac.kr).
}}

\maketitle

\begin{abstract}
    \indent We study hypergraph clustering in the weighted $d$-uniform hypergraph stochastic block model ($d$\textsf{-WHSBM}), where each edge consisting of $d$ nodes from the same community has higher expected weight than the edges consisting of nodes from different communities. We propose a new hypergraph clustering algorithm, called \textsf{CRTMLE}, and provide its performance guarantee under the $d$\textsf{-WHSBM} for general parameter regimes. We show that the proposed method achieves the order-wise optimal or the best existing results for approximately balanced community sizes. Moreover, our results settle the first recovery guarantees for growing number of clusters of unbalanced sizes. Involving theoretical analysis and empirical results, we demonstrate the robustness of our algorithm against the unbalancedness of community sizes or the presence of outlier nodes.
\end{abstract}

\section{Introduction}
\label{introduction}
\indent A hypergraph is an effective way to represent complex interactions among objects of interests. Different from classical graph modeling, where each edge connects only two nodes to model pairwise interactions, in hypergraphs an edge can connect more than two nodes to represent multi-way interactions among the nodes. Hypergraphs have been studied with diverse practical applications, such as clustering categorial databases \cite{GKR1998}, modeling folksonomies \cite{GZCN2009}, image segmentation \cite{ALZPKB2005}, and partitioning of circuit netlists in VLSI design \cite{KK2000}. \\
\indent In this paper, we study clustering problem in weighted uniform hypergraphs: Given a weighted hypergraph, our goal is to partition nodes into disjoint clusters so that within-cluster edges tend to have higher weights than cross-cluster edges. We propose an algorithm that recovers the hidden community structure from relatively sparse hypergraphs with growing number of unequal-sized communities, and analyze its statistical performance that either gives new consistency results for previously unknown parameter regimes or matches the best existing results. \\
\indent We focus on a generative random hypergraph model called the hypergraph stochastic block model, to evaluate hypergraph clustering algorithms. In graph clustering, the most widely studied model is the \textit{stochastic block model} (\textsf{SBM}) \cite{HLL1983}, also referred to as the \textit{planted partition model} \cite{CK2001}, where given an underlying partition $\Phi^*$ of $n$ nodes, a graph is generated such that two nodes in the same community are more likely to be adjacent than other pairs of nodes. We consider an extension of the standard \textsf{SBM} to weighted uniform hypergraphs, known as the \textit{weighted $d$-uniform hypergraph} \textsf{SBM} ($d$\textsf{-WHSBM}) \cite{ALS2018, GD2017-1}. We assume that all edges have the same size $d$. An edge is called $\Phi^*$-\textit{homogeneous} if it consists of $d$ nodes from the same community, and is called $\Phi^*$-\textit{heterogeneous} otherwise. In this model, a random weight is assigned from $[0, 1]$ independently to each edge such that $\Phi^*$-homogeneous edges tend to have higher weights with expectation $p_n$ than $\Phi^*$-heterogeneous edges, which have weights with expectation $q_n  < p_n$.

\subsection{Main Contributions}
\label{main_contributions}

We provide a hypergraph clustering algorithm based on the \textit{truncate-and-relax} strategy, called \textit{Convex Relaxation of Truncated Maximum Likelihood Estimator} (\textsf{CRTMLE}), which is motivated by \cite{KBG2017, KBG2018} under the unweighted hypergraph \textsf{SBM} with two equal-sized communities. \\
\indent Our algorithm can handle the high-dimensional case of $d$\textsf{-WHSBM} with hidden communities of order-wise unbalanced sizes. More precisely, our algorithm can operate in the $d$\textsf{-WHSBM} with parameters satisfying $(1)$ the number of communities $k$ may grow in $n$, and $(2)$ the order of community sizes can be different, \emph{i.e.}, $s_{\max} / s_{\min} = \omega(1)$, where $s_{\min}$ and $s_{\max}$ denote the minimum and maximum community sizes, respectively. As opposed to our general setup, most recent developments on efficient hypergraph clustering methods under variants of hypergraph \textsf{SBM} assume either the approximate balancedness of community sizes, \textit{i.e.}, $s_{\max} / s_{\min} = O(1)$, or the constant number of communities, \textit{i.e.}, $k = \Theta(1)$, for easiness of statistical analysis. \\
\indent Our main contribution is a statistical analysis of \textsf{CRTMLE} in general regimes for parameters $\left( p_n, q_n, s_{\min} \right)$, which are allowed to scale in $n$. Our main theorem shows that \textsf{CRTMLE} achieves the strong consistency (\textit{a.k.a.} the exact recovery, which means that all the nodes are clustered correctly \textit{w.h.p.}) provided that the density gap $p_n - q_n$ satisfies
\beq
    \label{eq1}
     p_n - q_n = \Omega \left( \frac{n^{\frac{d-2}{2}}\cdot \sqrt{ p_n \left( s_{\min} \log n + n \right)}}{s_{\min}^{d-1}} \right).
\eeq
Note that the condition \eqref{eq1} does not explicitly depend on the number of communities $k$ but only though $s_{\min}$. From this single condition, we can show that our algorithm achieves the order-wise optimal or the best known performance, which mainly assumes $s_{\max} / s_{\min} = \Theta(1)$ or $s_{\min} = s_{\max}$. Moreover, up to our knowledge, this is the first result showing a sufficient condition for exact recovery in the $d$\textsf{-WHSBM} with growing number of communities of order-wise unbalanced sizes. \\
\indent Our technical byproduct is the derivation of a sharp concentration bound on the spectral norm of a certain random matrix, called the \textit{similarity matrix}, which has dependency among entries. Most existing concentration results of random matrices are built upon the independence between entries. To resolve the dependency issues, we use the celebrated combinatorial argument developed by Friedman, Kahn and Szemer\'{e}di \cite{FKS1989, FO2005, CGJ2018}. Details can be found in Section \ref{concentration_bounds}. \\
\indent This paper is organized as follows. We provide an overview of related works in Section \ref{related_works} and introduce the $d$\textsf{-WHSBM} and formulate the hypergraph clustering problem under this model in Section \ref{problem_setup}. Section \ref{algorithm_description} presents our hypergraph clustering algorithm (\textsf{CRTMLE}), and Section \ref{main_results} provides its performance guarantee. In Section \ref{model_extensions}, we further discuss variants of the $d$\textsf{-WHSBM} and provide the performance guarantee of \textsf{CRTMLE} for the model variants. In Section \ref{empirical_results}, we provide some simulation results that demonstrate the robustness of \textsf{CRTMLE} against the unbalancedness of community sizes and the presence of outlier nodes. In this section, we also apply the proposed algorithm to subspace clustering \cite{ALZPKB2005}, to demonstrate the performance and robustness of CRMLE in real applications.
The proof of main theorem is provided in Section \ref{proof_thm4.1}, and other technical proofs are deferred to appendices. Section \ref{conclusion} is devoted for final remarks.

\subsection{Related Works}
\label{related_works}

\subsubsection{Graph Clustering}
\indent We first review existing results for graph clustering ($d = 2$) and compare them with our key condition \eqref{eq1}, which holds for any $d\geq 2$. The graph clustering has been extensively studied with full generality to find sufficient conditions of strong consistency with computationally-feasible algorithms. Under the \textsf{SBM} with $p_n = \Theta(1)$, $q_n = \Theta(1)$ and $\left| p_n - q_n \right| = \Theta(1)$, exact recovery can be solved efficiently provided that $s_{\min} = \Omega (\sqrt{n})$ by spectral clustering \cite{CCT2012} or convexified MLE \cite{OH2011, A2014, CSX2014}. We remark that this state-of-the-art result is also valid for variants of the \textsf{SBM} which allows semi-randomness \cite{FK2001, KV2006} or outlier nodes. From the key condition \eqref{eq1}, one can see that if $p_n = \Theta(1)$, $q_n = \Theta(1)$ and $\left| p_n - q_n \right| = \Theta(1)$, \textsf{CRTMLE} achieves the state-of-the-art result  $s_{\min} = \Omega \left( \sqrt{n} \right)$ of the exact recovery for general $d \geq 2$. \\ 
\indent When edge densities are in the form of $p_n = p \alpha_n$ and $q_n = q \alpha_n$ for some constants $p > q > 0$, and the number of communities $k$ may grow in $n$, the sparsity level $\alpha_n$ which allows the exact recovery in the \textsf{SBM} within polynomial time is known to be $\alpha_n = \Omega \left( \left( n + s_{\min} \log n \right) / s_{\min}^2 \right)$ \cite{CL2015}. The key condition \eqref{eq1} reads $\alpha_n = \Omega \left( \left( n^{d-2} \left( s_{\min} \log n + n \right) \right) / s_{\min}^{2d-2} \right)$, which coincides with the above result for the graph case. For the standard \textsf{SBM} with constant number of communities of approximately balanced sizes, various computationally-efficient methods achieve the order-wise optimal sparsity level $\alpha_n = \Omega \left( \log n / n \right)$ \cite{MNS2015, AS2015, ABH2015, HWX2016}, and our result also achieves this limit under the above setting.

\subsubsection{Hypergraph Clustering}

\indent We next provide review for the hypergraph clustering from three different perspectives: (a) formulation of a generative random hypergraph model, (b) characterization of information-theoretic thresholds, and (c) identification of computational limits of efficient algorithms.  \\

\noindent \textbf{Generative random hypergraph model.} \hspace{0.1in} One of the most widely studied random models for hypergraph is the \emph{hypergraph} \textsf{SBM}. Most of the previous works for hypergraph SBM assume either the constant number of communities or the balancedness of the community sizes. In \cite{KBG2017, KBG2018}, the hypergraph \textsf{SBM} with two equal-sized clusters is studied, which is also known as the \emph{hypergraph planted bisection model}. The weighted hypergraph case is probed in \cite{ALS2018} for the case where the number of communities is fixed and community sizes are approximately balanced. In \cite{CZ2020}, the dense regime (\emph{i.e.}, $p_n = \Theta(1)$ and $q_n = \Theta(1)$) for hyperedges is investigated with multiple equal-sized communities. \\
\indent In this work, our focus is on the weighted $d$-uniform hypergraph \textsf{SBM} with growing number of clusters without any particular assumptions on the balancedness of community sizes. In this model, we assume that homogeneous edges, which consist of $d$ nodes from the same community, tend to have higher weights with expectation $p_n$, while heterogeneous edges having weights with expectation $q_n < p_n$. \\
\indent We remark that there are several extensions available for $d$-\textsf{WHSBM}. In \cite{GD2014, GD2015-1, GD2017-2,CLW2018, CLW2019}, hypergraph planted partition model or hypergraph \textsf{SBM} is considered where the hyperedge probabilities depend on \emph{edge homogeneity}  of the composed nodes, \emph{i.e.}, the more concentrated groups of nodes in terms of assigned communities are connected with higher probability.
Also, the non-uniform hypergraph \textsf{SBM}, in which the sizes of hyperedges may vary, is investigated in \cite{GD2017-1}. Recently, \cite{KSX2019} proposed a  model, called the hypergraph degree-corrected \textsf{SBM}, to handle the degree heterogeneity.. \\

\noindent \textbf{Information-theoretic limits.} \hspace{0.1in} After establishing the underlying random model suitable for each application of hypergraphs, one should decide a target recovery type (\emph{a.k.a.}, consistency type of estimators). Typically, there are 3 recovery types (strong consistency, weak consistency, and detection) in estimating the ground-truth community assignment, which will be elaborated in Section \ref{problem_setup}. In \cite{ALS2018, GD2014, GD2015-1, GD2017-1, GD2017-2, KSX2019}, the weak consistency is studied. The strong consistency conditions are analyzed in \cite{ALS2018, KBG2017, KBG2018}. There are limited number of works on detection of communities in hypergraphs \cite{PZ2019, ALS2018}. In this study, we focus on the strong consistency in recovering the ground-truth communities. \\
\indent A sharp statistical limit for the strong consistency is established for the hypergraph planted bisection model in \cite{KBG2018}. The optimal minimax rates of the fraction of misclustering error are analyzed in \cite{CLW2018, CLW2019} in the hypergraph \textsf{SBM} which reflects the edge homogeneity. For the binary-edge case, \cite{ALS2018} shows that a spectral clustering method with local refinement procedure achieves the order-wise optimal limit for the strong consistency when the number of communities is constant and their sizes are approximately balanced. In Section \ref{comparison_literature}, we show this optimality can be achieved by \textsf{CRTMLE} in the same regime. As this order-wise threshold is achieved via polynomial-time methods, it is not only the (order-wise) information-theoretic limit, but also the computational limit. \\

\noindent\textbf{Efficient algorithms and computational limits.}  In order to establish the computational limit, a polynomial-time algorithm should be designed with which the target recovery type can be achieved. In comparison to graph clustering literature, strong consistency in hypergraphs has been studied mainly assuming balanced community structures to provide theoretical analysis on the proposed methods. Up to our knowledge, our result is the first one that provides a sufficient condition for the strong consistency that can be achieved within polynomial-time even for the case with growing number of communities of order-wise unbalanced sizes. \\ 
\indent Due to remarkable practical advantages in implementation and computational aspects, many recent works on hypergraph clustering are built upon the spectral clustering methodology. One noteworthy approach is to truncate the observed hypergraph down to a weighted graph, where the edge weight assigned to $\left\{ i, j \right\} \in \binom{[n]}{2}$ equals the number of hyperedges containing both $i$ and $j$. Then, the latent membership structure is inferred by applying the standard spectral clustering method to either the adjacency matrix of the truncated weighted graph \cite{ALS2018, CZ2020, GD2015-1, GD2017-1} or the hypergraph Laplacian matrix \cite{GD2017-2, CLW2018, CLW2019}. Another prominent approach is to conduct a \textit{higher-order singular value decomposition} (\textsf{HOSVD}) \cite{KB2009, LDV2000} on the adjacency tensor of the observed hypergraph and then run $k$-means clustering on the output matrix obtained by \textsf{HOSVD} \cite{GD2014, GD2015-2}. Recently, \cite{KSX2019} developed a new method to cope with the \textit{degree heterogeneity} in hypergraphs. \\
\indent Compared to spectral clustering, there are limited endeavors with convex relaxation approach for hypergraphs. In \cite{KBG2017, KBG2018}, an efficient algorithm is designed based on the \textit{truncate-and-relax} strategy which consists of two stages: (1) truncate an observed hypergraph down to a weighted graph, and (2) relax a combinatorial optimization problem on the truncated objective function. They derived its strong consistency result under the hypergraph \textsf{SBM} with two equal-sized communities. We extend it to the weighted case with multiple communities of unequal sizes. The reason that we consider the convex relaxation approach rather than the spectral clustering method in general parameter setup is that conventional spectral clustering methods are known to be sensitive to the unbalancedness of community sizes and the presence of outlier nodes in graphs \cite{CCT2012, C2004, LCX2018}. In Section \ref{empirical_results}, we manifest the robustness of \textsf{CRTMLE} against the unbalancedness of community sizes and the presence of outlier nodes via experimental results and compare it to other spectral methods.

\subsection{Notations}
\label{notations}
Let $\mathbf{M}_{i*}$ and $\mathbf{M}_{*j}$ denote the $i^{\textnormal{th}}$ row and the $j^{\textnormal{th}}$ column of $\mathbf{M} \in \mathbb{R}^{m \times n}$, respectively. For any vector $d = \left( d_1, \cdots, d_n \right) \in \mathbb{R}^n$, $\textnormal{diag}(d)$ denotes the $n \times n$ diagonal matrix with diagonal entries $d_1, \cdots, d_n$. For any positive integers $m$ and $n$, we denote by $\mathbf{1}_{m \times n}$ the $m \times n$ all-one matrix and $\mathbf{I}_{n}$ the $n \times n$ identity matrix. For any $n \times n$ real symmetric matrix $\mathbf{S}$, let $\lambda_{i}(\mathbf{S})$ denote the $i^{\textnormal{th}}$ largest eigenvalue of $\mathbf{S}$. For any $v \in \mathbb{R}^n$ and an integer $d \geq 2$, the $d$-fold tensor product of $v$, $v^{\otimes d} \in \mathbb{R}^{[n]^d}$, is given by $v^{\otimes d}(i_1, i_2, \cdots, i_d) := \prod_{k=1}^{d} v_{i_k}$ for every $i_1, i_2, \cdots, i_d \in [n]$. Naturally, the inner product of two real $n$-dimensional $d$-tensors $\mathbf{A}, \mathbf{B} \in \mathbb{R}^{[n]^d}$ is defined by $\left\langle \mathbf{A}, \mathbf{B} \right\rangle := \sum_{i_1 = 1}^{d}\sum_{i_2 = 1}^{d} \cdots \sum_{i_d = 1}^{d} \mathbf{A}(i_1, i_2, \cdots, i_d) \mathbf{B}(i_1, i_2, \cdots, i_d)$. \\
\indent For $n \in \mathbb{N}$, let $[n] := \left\{ 1, 2, \cdots, n \right\}$. Given a set $\mathcal{A}$ and a non-negative integer $m$, we set $\binom{\mathcal{A}}{m} := \left\{ \mathcal{B} \subseteq \mathcal{A} : |\mathcal{B}| = m \right\}$ and $\binom{\mathcal{A}}{\leq m} := \left\{ \mathcal{B} \subseteq \mathcal{A} : \left| \mathcal{B} \right| \leq m \right\} = \cup_{l=0}^{m} \binom{\mathcal{A}}{l}$.

\section{Problem Setup}
\label{problem_setup}

Let $\mathcal{V} := [n]$ be the set of $n$ vertices and $\mathcal{E} := \binom{[n]}{d}$ denote the set of all edges of size $d$ (\textit{a.k.a.}, \textit{$d$-regular edges}) for a fixed integer $d \geq 2$. Also, let $k$ be the number of communities that may depend on $n$ and $\mathcal{P}(n, k)$ denote the set of all partitions of $n$ nodes into $k$ communities. \\
\indent Any partition $\Phi : [n] \rightarrow [k]$ in $\mathcal{P}(n, k)$ can be characterized as a \textit{membership matrix} $\mathbf{Z}(\Phi) \in \left\{0, 1 \right\}^{n \times k}$ given by $\left[ \mathbf{Z}(\Phi) \right]_{ij} = 1$ if $j = \Phi(i)$ and $0$ otherwise. Let $\mathcal{Z}(n, k) := \left\{ \mathbf{Z}(\Phi) : \Phi \in \mathcal{P}(n, k) \right\}$  denote the set of all membership matrices corresponding to partitions in $\mathcal{P}(n, k)$. We say that a $d$-regular edge $e = \{i_1, i_2, \cdots, i_d\} \in \mathcal{E}$ is \textit{$\Phi$-homogeneous} if $\Phi(i_1) = \Phi(i_2) = \cdots = \Phi(i_d)$, and \textit{$\Phi$-heterogeneous} otherwise. A natural concept to characterize the homogeneity of $d$-regular edges with respect to the partition $\Phi$ is the \textit{cluster tensor} given by $\mathbf{T}(\Phi)(i_1, i_2, \cdots, i_d) = 1$ if $\Phi(i_1) = \Phi(i_2) = \cdots = \Phi(i_d)$, and $0$ otherwise. Note that $\mathbf{T}(\Phi)$ is symmetric and we have $\mathbf{T}(\Phi) = \sum_{j=1}^{k} \left[ \mathbf{Z}(\Phi) \right]_{*j}^{\otimes d}$ for every $\Phi \in \mathcal{P}(n, k)$. Also, let $\mathcal{T}(n, k) := \left\{ \mathbf{T}(\Phi) : \Phi \in \mathcal{P}(n, k) \right\}$ be the set of all cluster tensors corresponding to partitions $\Phi \in \mathcal{P}(n, k)$. We now formally define the weighted $d$-uniform hypergraph \textsf{SBM}, which has an abbreviation $d$\textsf{-WHSBM}.

\begin{defi} [The $d$\textsf{-WHSBM}] 
\label{defi2.1}
\textnormal{With parameters $n, k \in \mathbb{N}$, $p_n, q_n \in [0,1]$ and $\Phi^* \in \mathcal{P}(n, k)$, the \textit{weighted $d$-uniform hypergraph stochastic block model} is a generative random hypergraph model which samples a weighted $d$-uniform hypergraph $\mathcal{H} = \left( [n], \mathbf{W} = \left( W_e : e \in \mathcal{E} \right) \right)$ according to the following rule: a random weight $W_e \in [0,1]$ is assigned to each $d$-regular edge $e \in \mathcal{E}$ independently; we have $\mathbb{E}[W_e] = p_n$ if $e \in \mathcal{E}$ is a $\Phi^*$-homogeneous edge, and $\mathbb{E}[W_e] = q_n$ otherwise. This model is denoted by $d\textsf{-WHSBM}(n, k, p_n, q_n, \Phi^*)$ and the parameter $\Phi^* \in \mathcal{P}(n, k)$ is called the \textit{ground-truth partition} or the \textit{ground-truth community assignment}.
}
\end{defi}

\indent Observe that $d$\textsf{-WHSBM} does not specify the edge weight distribution, but only specifies their expectations. Following the standard \textsf{SBM} literature, we mainly focus on the case $p_n > q_n$ (assortative case) throughout this paper. The case $p_n < q_n$ (disassortative case) can be discussed by considering the \textit{complement hypergraph} $\overline{\mathcal{H}} := \left( [n], \overline{\mathbf{W}} \right)$ of the given hypergraph $\mathcal{H}$, where $\overline{\mathbf{W}} := \left( 1 - W_{e} : e \in \mathcal{E} \right)$. \\
\indent While our model looks similar to the one in \cite{ALS2018}, the number of communities $k$ may grow in $n$ (\emph{a.k.a.}, the high-dimensional regime  \cite{CX2014, CX2016}) in our model. We denote by $\mathbf{Z}^* := \mathbf{Z}(\Phi^*)$ and $\mathbf{T}^* := \mathbf{T}(\Phi^*)$ the ground-truth membership matrix and the ground-truth cluster tensor, respectively. Throughout this paper, we write $C_{a}^* := \left( \Phi^* \right)^{-1}(a)$, $a \in [k]$, to denote the $a^{\textnormal{th}}$ ground-truth community. We also denote by $s_a := | C_{a}^* |$, $s_{\textnormal{min}} := \min \left\{ s_a : a \in [k] \right\}$ and $s_{\textnormal{max}} := \max \left\{ s_a : a \in [k] \right\}$ the size of the $a^{\textnormal{th}}$ community, the minimum and the maximum sizes of communities, respectively. We remark that $d$\textsf{-WHSBM} extends the graph case ($d = 2$) to the $d$-uniform hypergraph setting for general $d \geq 2$ \cite{JL2015, XJL2017}.

\medskip\noindent
{\textbf{Consistency types of estimators.} \hspace{0.1in}}
Given a weighted random $d$-uniform hypergraph $\mathcal{H} = \left( [n], \mathbf{W} \right)$, we want to recover the ground-truth community assignment $\Phi^*$ \textit{up to a permutation}. To be precise, given any estimator $ \hat{\Phi} = \hat{\Phi}(\mathbf{W}) : [n] \rightarrow [k]$, we define the \textit{fraction of misclustering error} of $\hat{\Phi}$ by
\begin{equation*}
    \textnormal{err}(\hat{\Phi}, \Phi^*) := \frac{1}{n} \min_{\pi \in \mathcal{S}_{k}} \left| \left\{ i \in [n] : \Phi^*(i) \neq \pi ( \hat{\Phi}(i) ) \right\} \right|,
\end{equation*}
where $\mathcal{S}_{k}$ denotes the symmetric group of degree $k$. Then, we say that an estimator $\hat{\Phi} = \hat{\Phi}(\mathbf{W}) : [n] \to [k]$ is
\begin{enumerate}
    \item \emph{strongly consistent} (\emph{a.k.a.}, solving the \emph{exact recovery}) if $\lim_{n \to \infty} \mathbb{P} \{ \textnormal{err}(\hat{\Phi}, \Phi^*) = 0 \} = 1$.
    \item \emph{weakly consistent} (\emph{a.k.a.}, solving the \emph{almost exact recovery}) if $\textnormal{err}( \hat{\Phi}, \Phi^*) \to 0$ in probability as $n \to \infty$.
    \item solving the \emph{detection} if there exists a positive real number $\epsilon > 0$ such that $\lim_{n \to \infty} \mathbb{P} \left\{ \textnormal{err}(\hat{\Phi}, \Phi^*) \geq \frac{1}{k} + \epsilon \right\} = 1$.
\end{enumerate}
Throughout this paper, we will focus on the strong consistency of estimators (\textit{a.k.a.} the \textit{exact recovery} \cite{A2018, DKMZ2011}).

\begin{table}[t]
    \centering
    \caption{Summary of Notations and Descriptions}
    \label{tab:table1}
    \resizebox{\columnwidth}{!}{\begin{tabular}{|c|l|}
        \hline
        \thead{\textbf{Notation}} & \thead{\textbf{Description}} \\ \hline  
        $\mathcal{H} = \left( [n], \mathbf{W} \right)$ & A weighted $d$-uniform hypergraph \\ \hline
        $\mathcal{E}$ & The set of all $d$-regular edges of $n$ vertices \\ \hline
        $\Phi : [n] \rightarrow [k]$ & A partition of $n$ nodes into $k$ communities \\ \hline
        $\mathbf{Z}(\Phi) \in \left\{0, 1 \right\}^{n \times k}$ & The membership matrix corresponding to $\Phi$ \\ \hline
        $\mathbf{T}(\Phi) = \sum_{j=1}^{k} \left[ \mathbf{Z}(\Phi) \right]_{*j}^{\otimes d}$ & The cluster tensor corresponding to $\Phi$ \\ \hline
        $\mathbf{X}(\Phi):= \mathbf{Z}(\Phi) \left[ \mathbf{Z}(\Phi) \right]^{\top} $ & The cluster matrix corresponding to $\Phi$ \\ \hline
        $\Phi^*, \mathbf{Z}^*, \mathbf{T}^*, \mathbf{X}^*$ & The ground-truth quantities  \\ \hline
        $C_{a}^*:= \left( \Phi^* \right)^{-1}(a), s_a:=|C_{a}^*|$ & The $a^{\textnormal{th}}$ ground-truth community and its size \\ \hline
        $s_{\min}, s_{\max}$ & The minimum and maximum sizes of clusters \\ \hline
    \end{tabular}}
\end{table}

\indent The set of notations introduced in this section is summarized in Table \ref{tab:table1}.

\section{Algorithm Description}
\label{algorithm_description}

Our proposed algorithm, \textit{Convex Relaxation of Truncated Maximum Likelihood Estimator} (\textsf{CRTMLE}), consists of three main stages: (1) truncation of maximum likelihood estimator (MLE), (2) semi-definite program (SDP) relaxation and (3) explicit clustering via approximate $k$-medoids clustering method. We first explain those steps to motivate the algorithm and then provide the complete algorithm.

\subsubsection{Truncation of maximum likelihood estimator}
Given a sample $\mathcal{H} = \left( [n], \mathbf{W} \right)$ drawn by $d\textsf{-WHSBM}(n, k, p_n, q_n, \Phi^*)$, we consider the MLE $\hat{\mathbf{T}}_{\textnormal{MLE}}(\mathbf{W})$ of the \textit{ground-truth cluster tensor} $\mathbf{T}^*$. We analyze the MLE for the \textit{binary-valued edge weight case} and later show that the algorithm developed for the binary-edge case achieves the strong consistency guarantee even for the general weighted case. \\
\indent The log-likelihood function of observing the binary-valued weighted hypergraph $\mathbf{W} = \left( W_e : e \in \mathcal{E} \right)$ given a cluster tensor $\mathbf{T} \in \mathcal{T}(n, k)$ is

\beq
    \begin{split}
        \label{eq2}
        &\ \log \mathbb{P} \{\mathbf{W} | \mathbf{T}\} \\
        =&\ \log \prod_{e \in \mathcal{E}} \left[ \left( \frac{p_n (1-q_n)}{q_n (1-p_n)} \right)^{W_e T_e} \left( \frac{1-p_n}{1-q_n} \right)^{T_e} \times \right.\\
        &\qquad\qquad \left. q_{n}^{W_e} (1-q_n)^{1-W_e} \right] \\
        =&\ \log \left( \frac{p_n (1-q_n)}{q_n (1-p_n)} \right) \sum_{e \in \mathcal{E}} W_e T_e - \log \left( \frac{1-q_n}{1-p_n} \right) \sum_{e \in \mathcal{E}} T_e \\
        &+ \left( \textnormal{constant terms of $\mathbf{T}$} \right) \\
        =&\ \frac{1}{d!} \left[ \log \left( \frac{p_n (1-q_n)}{q_n (1-p_n)} \right) \langle \mathbf{W}, \mathbf{T} \rangle - \log \left( \frac{1-q_n}{1-p_n} \right) \langle \mathbbm{1}_{\mathcal{E}}, \mathbf{T} \rangle \right] \\
        &+ \left( \textnormal{constant terms of $\mathbf{T}$} \right),
    \end{split}
\eeq
where $T_{e} := \mathbf{T}(i_1, i_2, \cdots, i_d)$ for every $d$-regular edge $e = \left\{ i_1, i_2, \cdots, i_d \right\} \in \mathcal{E}$ (this convention is well-defined since $\mathbf{T}$ is a symmetric $d$-tensor.), and $\mathbbm{1}_{\mathcal{E}} \in \mathbb{R}^{[n]^d}$ denotes the indicator tensor of the $d$-regular edge set $\mathcal{E}$, \emph{i.e.}, $\mathbbm{1}_{\mathcal{E}}(i_1, i_2, \cdots, i_d) := 1$ if $i_1, i_2, \cdots, i_d \in [n]$ are all distinct and $0$ otherwise. Here, we may view $\mathbf{W}$ as a symmetric $n$-dimensional $d$-tensor given by $\mathbf{W}(i_1, i_2, \cdots, i_d) := W_{\left\{ i_1, i_2, \cdots, i_d \right\}}$ if $\mathbbm{1}_{\mathcal{E}}(i_1, i_2, \cdots, i_d) = 1$, and $0$ otherwise. The assumption $p_n > q_n$ gives $\hat{\mathbf{T}}_{\textnormal{MLE}}(\mathbf{W}) \in$
\beq
    \label{eq3}
    \arg \max \left\{ \langle \mathbf{W}, \mathbf{T} \rangle - \mu \langle \mathbbm{1}_{\mathcal{E}}, \mathbf{T} \rangle : \mathbf{T} \in \mathcal{T}(n, k) \right\},
\eeq
where $\mu = \mu(p_n, q_n) := \frac{\log(1-q_n) - \log(1-p_n)}{\log p_n + \log(1-q_n) - \log q_n - \log(1-p_n)} > 0$. Intuitively, the optimization problem \eqref{eq3} seeks for a cluster tensor $\mathbf{T} \in \mathcal{T}(n, k)$ which maximizes a \emph{penalized correlation} with the observed data $\mathbf{W}$. Note that the regularization term plays a role in balancing the number of homogeneous edges and heterogeneous edges in order to avoid the circumstance of having too many edges aligned with groups of nodes belonging to the same community. \\
\indent For each cluster tensor $\mathbf{T}$, there is a corresponding membership matrix $\mathbf{Z} \in \mathcal{Z}(n, k)$ such that $\mathbf{T} = \sum_{j=1}^{k} \left( \mathbf{Z}_{*j} \right)^{\otimes d}$. By defining $\mathbf{Y} := 2\mathbf{Z} - \mathbf{1}_{n \times k}$ and $\mathcal{Y}(n, k) := \{ 2 \mathbf{Z} - \mathbf{1}_{n \times k} \in \left\{ \pm 1 \right\}^{n \times k} : \mathbf{Z} \in \mathcal{Z}(n, k) \}$, we can represent the MLE \eqref{eq3} in terms of $\mathbf{Y}$: $\hat{\mathbf{Y}}_{\textnormal{MLE}}(\mathbf{W}) \in$
\beq
    \label{eq4}
    \begin{split}
    \arg \max \left\{ \left\langle \mathbf{W} - \mu \mathbbm{1}_{\mathcal{E}}, \sum_{j=1}^{k} \left( \mathbf{Y}_{* j} + \mathbf{1}_n \right)^{\otimes d} \right\rangle : \mathbf{Y} \in \mathcal{Y}(n, k) \right\},
    \end{split}
\eeq
where $\mathbf{1}_n \in \mathbb{R}^n$ is the $n$-dimensional all-one vector. By utilizing the expansion $\left[ \left( \mathbf{Y}_{*j} + \mathbf{1}_{n} \right)^{\otimes d} \right]_{e} = \prod_{i \in e} \left( \mathbf{Y}_{ij} + 1 \right) = \sum_{I \subseteq e} \left( \prod_{i \in I} \mathbf{Y}_{ij} \right)$ for every $d$-regular edge $e \in \mathcal{E}$, one can observe that for every $\mathbf{Y} \in \mathcal{Y}(n, k)$, we may deduce that the objective function $f_{\mathbf{W}} : \mathcal{Y}(n, k) \to \mathbb{R}$ of \eqref{eq4} is given by
\begin{equation*}
    \begin{split}
        f_{\mathbf{W}}(\mathbf{Y}) 
        &\overset{\textnormal{(a)}}{=} d! \sum_{j=1}^{k}\sum_{e \in \mathcal{E}} \sum_{I \subseteq e} \left (W_e - \mu \right) \left( \prod_{i \in I} \mathbf{Y}_{ij} \right) \\
        &\overset{\textnormal{(b)}}{=} d! \sum_{j=1}^{k} \sum_{I \in \binom{[n]}{d}} \sum_{e \in \mathcal{E} : I \subseteq e} \left( W_e - \mu \right) \left( \prod_{i \in I} \mathbf{Y}_{ij} \right) \\
        &= d! \sum_{I \in \binom{[n]}{\leq d}} \left[ \sum_{e \in \mathcal{E} : I \subseteq e} \left( W_e - \mu \right) \right] \left[ \sum_{j=1}^{k} \left( \prod_{i \in I} \mathbf{Y}_{ij} \right) \right],
    \end{split}
\end{equation*}
where (a) is due to the symmetry of  $\mathbf{W}$ and $\left( \mathbf{Y}_{*j} + \mathbf{1}_n \right)^{\otimes d}$ and (b) is simply from interchanging the order of summation. Here, we define polynomials in an indeterminate $\mathbf{Y}$ by
\begin{equation*}
    (p_l)_{\mathbf{W}}(\mathbf{Y}) := \sum_{I \in \binom{[n]}{l}} \left[ \sum_{e \in \mathcal{E} : I \subseteq e} \left( W_e - \mu \right) \right] \left[ \sum_{j=1}^{k} \left( \prod_{i \in I} \mathbf{Y}_{ij} \right) \right]
\end{equation*}
for $l \in \left\{ 0, 1, \cdots, d \right\}$. Some straightforward calculations yield
\begin{equation*}
    \begin{split}
        &(p_0)_{\mathbf{W}}(\mathbf{Y}) = \frac{k}{d!} \langle \mathbf{W} - \mu \mathbbm{1}_{\mathcal{E}}, \mathbbm{1}_{\mathcal{E}} \rangle,\\
        &(p_1)_{\mathbf{W}}(\mathbf{Y}) = \frac{d(2-k)}{d!} \langle \mathbf{W} - \mu \mathbbm{1}_{\mathcal{E}}, \mathbbm{1}_{\mathcal{E}} \rangle, \\
        & (p_2)_{\mathbf{W}}(\mathbf{Y})= \\
        &\quad \sum_{1 \leq i < j \leq n} \left[ \left( \sum_{e \in \mathcal{E} : \{ i, j \} \subseteq e} W_e \right) - \mu \binom{n-2}{d-2} \right] \left[ \mathbf{Y}\mathbf{Y}^{\top} \right]_{ij},
    \end{split}
\end{equation*}
and thereby it follows that
\beq
    \label{eq5}
    \begin{split}
        f_{\mathbf{W}}(\mathbf{Y}) =& \left( k + 2d - dk \right) \langle \mathbf{W} - \mu \mathbbm{1}_{\mathcal{E}}, \mathbbm{1}_{\mathcal{E}} \rangle + d! (p_2)_{\mathbf{W}}(\mathbf{Y}) \\
        &+ \left( \textnormal{higher-order terms of } \mathbf{Y} \right).
    \end{split}
\eeq
\indent Instead of evaluating the maximum of the high-order polynomial $f_{\mathbf{W}}(\mathbf{Y})$ over $\mathbf{Y} \in \mathcal{Y}(n, k)$, we first approximate \eqref{eq5} by truncating terms of order higher than 2 and compute the maximum of the truncated polynomial over $\mathbf{Y} \in \mathcal{Y}(n, k)$. This computation is called the \textit{truncated maximum likelihood estimation} and  the corresponding estimator is denoted by $\hat{\mathbf{Y}}_{\textnormal{trunc}}$ given by $\hat{\mathbf{Y}}_{\textnormal{trunc}}(\mathbf{W}) \in \arg \max \left\{ (p_2)_{\mathbf{W}}(\mathbf{Y}) : \mathbf{Y} \in \mathcal{Y}(n, k) \right\}.$ It is clear that $\hat{\mathbf{Y}}_{\textnormal{trunc}}(\mathbf{W}) = 2 \hat{\mathbf{Z}}_{\textnormal{trunc}}(\mathbf{W}) - \mathbf{1}_{n \times k}$, where
\beq
    \label{eq6}
    \hat{\mathbf{Z}}_{\textnormal{trunc}}(\mathbf{W}) \in \arg \max \left\{ (p_2)_{\mathbf{W}}(\mathbf{Z}) : \mathbf{Z} \in \mathcal{Z}(n, k) \right\}.
\eeq

\noindent This degree-2 truncation strategy has been considered in \cite{KBG2017, KBG2018}, where the truncated MLE for the hypergraph \textsf{SBM} was derived and analyzed for two balanced clusters. We analyze this estimator for general parameter regimes. The coefficients of the polynomial $(p_2)_{\mathbf{W}}(\mathbf{Y})$ stimulates us to consider the truncation of the weighted hypergraph $\mathcal{H} = \left( [n], \mathbf{W} \right)$ down to a weighted graph whose adjacency matrix is referred to as the \emph{similarity matrix} \cite{ALS2018, KBG2018, FL1996, CZ2020}:

\begin{defi} [Similarity Matrix] 
\label{defi3.1}
\normalfont{ 
\indent The \textit{similarity matrix} $\mathbf{A}$ of a weighted $d$-uniform hypergraph $\mathcal{H} = \left( [n], \mathbf{W} \right)$ is an $n \times n$ real symmetric matrix with entries $\mathbf{A}_{ij} := \sum_{e \in \mathcal{E} : \{i, j\} \subseteq e} W_e$ if $i \neq j$; and $\mathbf{A}_{ij} := 0$ otherwise.
}
\end{defi}

From $(p_2)_{\mathbf{W}}(\mathbf{Z}) = \frac{1}{2} \langle \mathbf{A} - \mu \binom{n-2}{d-2} \mathbf{1}_{n \times n}, \mathbf{Z}\mathbf{Z}^{\top} \rangle + \frac{n}{2} \mu \binom{n-2}{d-2}$, we arrive at the following equivalent formulation of \eqref{eq6}:
\beq
    \begin{split}
        \label{eq7}
        &\hat{\mathbf{Z}}_{\textnormal{trunc}}(\mathbf{W}) \in \\
        &\arg \max \left\{ \left\langle \mathbf{A} - \mu \binom{n-2}{d-2} \mathbf{1}_{n \times n}, \mathbf{Z} \mathbf{Z}^{\top} \right\rangle : \mathbf{Z} \in \mathcal{Z}(n, k) \right\}.
    \end{split}
\eeq
Note that the program \eqref{eq7} is non-convex and computationally infeasible since the feasible set $\mathcal{Z}(n, k)$ is discrete, non-convex and exponentially large as $\left| \mathcal{Z}(n, k) \right| = \left| \mathcal{P}(n, k) \right| = \Omega(e^n)$.

\subsubsection{Convex relaxation of truncated MLE}
To derive a convex relaxation of \eqref{eq7}, it is more convenient to recast the problem as the following form:
\beq
    \begin{split}
        \label{eq8}
        \max_{\mathbf{X} \in \mathcal{X}(n, k)} \ &\left\langle \mathbf{A} - \mu \binom{n-2}{d-2} \mathbf{1}_{n \times n}, \mathbf{X} \right\rangle,
    \end{split}
\eeq
where $\mathcal{X}(n, k) := \left\{ \mathbf{Z} \mathbf{Z}^{\top} : \mathbf{Z} \in \mathcal{Z}(n, k) \right\}$. Let $\mathbf{X}(\Phi) := \mathbf{Z}(\Phi) \left[ \mathbf{Z}(\Phi) \right]^{\top} \in \mathcal{X}(n, k)$ denote the \textit{cluster matrix} corresponding to $\Phi \in \mathcal{P}(n, k)$. We may observe that any $\mathbf{X} \in \mathcal{X}(n, k)$ satisfies the following convex properties: $(1)$ all entries of $\mathbf{X}$ lie in $[0, 1]$, $(2)$ $\textnormal{Trace}(\mathbf{X}) = n$, and $(3)$ $\mathbf{X}$ is positive definite. By relaxing the non-convex constraint in \eqref{eq8}, we obtain an SDP given by:  
\beq
    \begin{split}
        \label{eq9}
        \max_{\mathbf{X} \in \mathbb{R}^{n \times n}} \ &\langle \mathbf{A} - \lambda \mathbf{1}_{n \times n}, \mathbf{X} \rangle \\
        \textnormal{subject to } &\mathbf{X} \succeq \mathbf{O};\ \langle \mathbf{I}_n, \mathbf{X} \rangle = n; \\
        &0 \leq \mathbf{X}_{ij} \leq 1,\ \forall i, j \in [n],
    \end{split}
\eeq
where $\mathbf{O}$ is the $n \times n$ all-zero matrix. The tuning parameter $\lambda \geq 0$, which substitutes the coefficient $\mu \binom{n-2}{d-2}$, must be specified. One can think of the tuning parameter $\lambda$ as a regularization parameter that controls the \textit{sparseness} of $\mathbf{X}$ since $\langle \mathbf{1}_{n \times n}, \mathbf{X} \rangle = \|\mathbf{X}\|_{1}$, which is immediate from $\mathbf{X}_{ij} \geq 0$, $\forall i, j \in [n]$. An optimal solution $\hat{\mathbf{X}}_{\textnormal{SDP}}(\mathbf{W})$ to the SDP \eqref{eq9} will play a role as an estimator of the \textit{ground-truth cluster matrix} $\mathbf{X}^* := \mathbf{Z}^* \left( \mathbf{Z}^* \right)^{\top} \in \mathcal{X}(n, k)$.

\begin{rmk}[Tuning parameter $\lambda$]
\label{remark_tuning}
\textnormal{The proper choice of the tuning parameter $\lambda$ is important in the performance of SDP \eqref{eq9}. As will be shown in main results, the tuning parameter $\lambda$ should be chosen to lie between the minimum within-cluster similarity and the maximum cross-cluster similarity (which are specified in Section \ref{performance_analysis}) to guarantee the exact recovery. The parameter $\lambda$ specifies the resolution of clustering algorithm: a higher $\lambda$ tends to detect smaller clusters with similarity (in the similarity matrix $\mathbf{A}$) larger than $\lambda$. So, varying $\lambda$ results in different solutions with cluster resolutions determined by $\lambda$. For this reason, it is not generally possible to determine a unique choice of $\lambda$ from the data. Similar phenomenon has been known for an SDP for graph clustering \cite{CSX2014}. If the community sizes are all equal, on the other hand, it is possible to determine a proper choice of $\lambda$ in a completely data-driven way with a theoretical guarantee (see Section \ref{estimation_lambda} for details).}
\end{rmk}

\begin{rmk}
\label{alternative_SDP}
\textnormal{
Instead of  \eqref{eq9}, we may consider the alternative SDP below:
\begin{equation}
    \label{eq10}
    \begin{split}
        \max_{\mathbf{X} \in \mathbb{R}^{n \times n}} \ &\langle \mathbf{A}, \mathbf{X} \rangle \\
        \textnormal{subject to } &\mathbf{X} \succeq \mathbf{0};\ 0 \leq \mathbf{X}_{ij} \leq 1,\ \forall i, j \in [n]; \\
        &\langle \mathbf{I}_{n}, \mathbf{X} \rangle = n;\ \langle \mathbf{1}_{n \times n}, \mathbf{X} \rangle = \langle \mathbf{1}_{n \times n}, \mathbf{X}^* \rangle.
    \end{split}
\end{equation}
We remark that the program \eqref{eq9} can be viewed as a \textit{penalized form} of \eqref{eq10}, obtained by removing the constraint on the term $\langle \mathbf{1}_{n \times n}, \mathbf{X} \rangle$ but instead penalizing it. Since $\mathbf{X}_{ij}^* = 1$ if and only if both $i$ and $j$ belong to the same ground-truth community, one has $\langle \mathbf{1}_{n \times n}, \mathbf{X}^* \rangle = \sum_{i=1}^{n} \sum_{j=1}^{n} \mathbf{X}_{ij}^* = \sum_{a=1}^{k} s_{a}^2$. Thus, the alternative SDP \eqref{eq10} does not require parameter tuning for $\lambda$ compared to \eqref{eq9}, but instead it requires the exact knowledge of the sum of squares of the community sizes $\sum_{a=1}^{k} s_{a}^2$. This requisite may be unrealistic to assume in practical applications, however it becomes more reasonable whenever clusters are all equal-sized. In this section, we focus on the penalized SDP \eqref{eq9}, and we turn to the alternative SDP \eqref{eq10} later when we allow the existence of outlier nodes, which refer to nodes that belong to no community, for the balanced case.
}
\end{rmk}

\subsubsection{Explicit clustering via approximate $k$-medoids clustering}
We next present a method to extract an explicit clustering from the solution $\hat{\mathbf{X}}_{\textnormal{SDP}} = \hat{\mathbf{X}}_{\textnormal{SDP}}(\mathbf{W})$ of the SDP \eqref{eq9}. The basic idea is to use $k$-medoids clustering for the $n$ row vectors of $\hat{\mathbf{X}}_{\textnormal{SDP}}$.
To explain the algorithm details, we first review \emph{$k$-medoids clustering problem}. Let $\mathbb{X} := \left\{ x_1, x_2, \cdots, x_n \right\}$ be the given $n$ input data points that lie in the ambient Euclidean space $\mathbb{R}^p$, and $\mathbf{X} \in \mathbb{R}^{n \times p}$ be a matrix with $\mathbf{X}_{i*} = x_i$, $\forall i \in [n]$. The $k$-medoids clustering problem searches for a clustering assignment of these $n$ data points and the corresponding centers of $k$ clusters $\left\{ v_1, v_2, \cdots, v_k \right\} \subseteq \mathbb{X}$ such that the sum of $l_1$-norms of each data point to its cluster center is minimized. We can formalize this problem as below: 
\beq
    \label{eq11}
    \begin{split}
        \min_{\Phi, v_1, \cdots, v_k} \ &\sum_{a=1}^{k} \left[ \sum_{i \in \Phi^{-1}(a)} \left\| x_i - v_a \right\|_{1} \right] \\
        \textnormal{subject to } &\Phi \in \mathcal{P}(n, k); \\
        &v_1, v_2, \cdots, v_k \in \mathbb{X}.
    \end{split}
\eeq
Representing the cluster centers $\left\{ v_1, v_2, \cdots, v_k \right\}$ as row vectors of a matrix $\mathbf{V} \in \mathbb{R}^{k \times p}$, the problem \eqref{eq11} can be re-written as the following compact form:
\beq
    \label{eq12}
    \begin{split}
        \min_{\mathbf{Z}, \mathbf{V}} \ &\left\| \mathbf{Z}\mathbf{V} - \mathbf{X} \right\|_{1} \\
        \textnormal{subject to } &\mathbf{Z} \in \mathcal{Z}(n, k),\ \mathbf{V} \in \mathbb{R}^{k \times p}; \\
        &\textnormal{Rows}(\mathbf{V}) \subseteq \textnormal{Rows}(\mathbf{X}),
    \end{split}
\eeq
where $\textnormal{Rows}(\mathbf{M})$ denotes the set of row vectors of a matrix $\mathbf{M}$, and $\left\| \mathbf{M} \right\|_{1}$ is the sum of absolute values of all entries of $\mathbf{M}$. A common method to solve the $k$-medoids problem \eqref{eq11} is the \emph{$k$-medoids clustering algorithm}, which approximately minimizes the objective function of \eqref{eq11} by alternately minimizing over $\Phi$ and $v_1, v_2, \cdots, v_k$. We may extract a clustering from $\hat{\mathbf{X}}_{\textnormal{SDP}}$ by applying the $k$-medoids clustering method on its row vectors. However, it has a crucial shortcoming in computational aspect since finding the medoid of given dataset is computationally hard in general. More generally, computing the exact optimizer $( \tilde{\mathbf{Z}}, \tilde{\mathbf{V}})$ of \eqref{eq12} is known to be NP-hard. \\
\indent To address this issue, \cite{CGTS2002} proposed an efficient algorithm to solve the $k$-medoids problem \emph{approximately}. This algorithm produces an output $(\hat{\mathbf{Z}}, \hat{\mathbf{V}})$ feasible to \eqref{eq12} within polynomial-time such that
\begin{equation*}
    \left\| \hat{\mathbf{Z}} \hat{\mathbf{V}} - \mathbf{X} \right\|_{1} \leq \frac{20}{3} \left\| \tilde{\mathbf{Z}} \tilde{\mathbf{V}} - \mathbf{X} \right\|_{1}.
\end{equation*}
From this output $(\hat{\mathbf{Z}}, \hat{\mathbf{V}})$, we can extract a community assignment $\hat{\Phi}$ by letting $\hat{\Phi}(i)$ be the unique non-zero coordinate of $\hat{\mathbf{Z}}_{i*}$. This efficient clustering extraction method was originally designed in \cite{FC2018, CLX2018}, and is called the \emph{approximate $k$-medoids clustering algorithm}. See \emph{Algorithm 1} in \cite{FC2018} for the detailed procedure. Moreover, the authors provide an error bound of the output of their method. See \emph{Proposition 3} therein: Given any estimator $\hat{\mathbf{X}} = \hat{\mathbf{X}}(\mathbf{W})$ of $\mathbf{X}^*$, let $\hat{\Phi}$ be the output of the approximate $k$-medoids clustering on $\hat{\mathbf{X}}$. Then, it satisfies
\begin{equation*}
    \textnormal{err}(\hat{\Phi}, \Phi^*) \leq \frac{86}{3} \frac{\| \hat{\mathbf{X}} - \mathbf{X}^*\|_{1}}{\| \mathbf{X}^* \|_{1}}.
\end{equation*}
This bound implies $\hat{\Phi}_{\textnormal{SDP}} = \Phi^*$ whenever $\hat{\mathbf{X}}_{\textnormal{SDP}} = \mathbf{X}^*$, where $\hat{\Phi}_{\textnormal{SDP}}$ denotes the output of \textsf{CRTMLE}. Furthermore, there have been some recent breakthroughs in randomized algorithms for finding the medoid of the given dataset by converting the medoid problem to a multi-armed bandit statistical inference problem \cite{BT2019, TZMTPS2020}. Applying them enables faster implementation of the $k$-medoids clustering algorithm.

\begin{algorithm}[tb]
	\caption{\textsf{CRTMLE}: Convex Relaxation of Truncated MLE}
	\label{algorithm1}

\begin{algorithmic}[1]
\STATE \textbf{Data}: A weighted $d$-uniform hypergraph $\mathcal{H} = \left( [n], \mathbf{W} \right)$, a tuning parameter $\lambda > 0$.
\STATE Compute the similarity matrix $\mathbf{A} \in \mathbb{R}^{n \times n}$ of $\mathcal{H}$.
\STATE Solve the SDP \eqref{eq9} with $\mathbf{A}$. Let $\hat{\mathbf{X}}_{\textnormal{SDP}} = \hat{\mathbf{X}}_{\textnormal{SDP}}(\mathbf{W})$ be an optimal solution.
\STATE Employ the approximate $k$-medoids clustering (\textit{Algorithm 1} in \cite{FC2018}) on $\hat{\mathbf{X}}_{\textnormal{SDP}}$ for extraction of an explicit community assignment, $\hat{\Phi}_{\textnormal{SDP}}(\mathbf{W}) : [n] \rightarrow [k]$.
\STATE \textbf{Output}: The community assignment $\hat{\Phi}_{\textnormal{SDP}} = \hat{\Phi}_{\textnormal{SDP}}(\mathbf{W})$.
\end{algorithmic}

\end{algorithm}

\begin{rmk}[Computational Tractability of \textsf{CRTMLE}]
\label{remark_CRTMLE}
\textnormal{The time complexity of step 2 in \textsf{CRTMLE} (see Alg. \ref{algorithm1} for the detailed procedure) is $O \left( 2 \binom{d}{2} |\mathcal{E}| \right) = O(n^d)$ since each $d$-regular edge $e \in \mathcal{E}$ appears $2 \binom{d}{2}$ times during the construction of $\mathbf{A}$. Also we note that SDPs are solvable efficiently either by the \textit{interior point method} \cite{A1995} or the \textit{alternating direction method of multipliers} \cite{BPCPE2011}. Thus, step 3 in Alg. \ref{algorithm1} can be done within polynomial-time. Finally, as we discussed in Section \ref{algorithm_description}-3), the approximate $k$-medoids clustering (step 4 in Alg. \ref{algorithm1}) operates within polynomial-time. Thus, \textsf{CRTMLE} is a polynomial-time algorithm, \emph{i.e.}, computationally tractable regardless of its success or failure in the exact recovery.
}
\end{rmk}

\section{Main Results}
\label{main_results}

In this section, we analyze the statistical performance of our proposed algorithm and compare it with existing works (Section \ref{performance_analysis} and \ref{comparison_literature}). Finally, we discuss how we can estimate the tuning parameter $\lambda$ required in the SDP \eqref{eq9} under the balanced case (Section \ref{estimation_lambda}).

\subsection{Performance Analysis of \textnormal{\textsf{CRTMLE}}} 
\label{performance_analysis}

While our main algorithm (\textsf{CRTMLE}) is derived from the MLE for the binary-valued edge case of $d$\textsf{-WHSBM}, we study its performance guarantee for the general case. We aim to characterize a sufficient condition for the strong consistency of \textsf{CRTMLE} under the $d$\textsf{-WHSBM}. \\
\indent Consider any two off-diagonal entries $\mathbf{A}_{ij}$ and $\mathbf{A}_{i'j'}$ of the similarity matrix with $\Phi^*(i) = \Phi^*(i')$ and $\Phi^*(j) = \Phi^*(j')$. Then, one can see that $\mathbb{E}[\mathbf{A}_{ij}] = \mathbb{E}[\mathbf{A}_{i'j'}]$. Consequently, we can define a $k \times k$ real symmetric matrix $\Delta$ whose entries are given by $\Delta_{ab} := \mathbb{E}[\mathbf{A}_{ij}]$, where $i \in C_{a}^*$ and $j \in C_{b}^*$, for some $i \neq j$. We can compute the entries of $\Delta$ explicitly as
\begin{equation*}
    \begin{split}
        \Delta_{aa} &= \binom{s_{a}-2}{d-2} \left( p_n - q_n \right) + \binom{n-2}{d-2} q_n, \ \forall a \in [k]; \\
        \Delta_{ab} &= \binom{n-2}{d-2} q_n, \ \forall a \neq b \textnormal{ in } [k].
    \end{split}
\end{equation*}
One can see that the diagonal entries of $\Delta$ are strictly larger than its off-diagonal entries in the assortative $d$\textsf{-WHSBM}. The diagonal entries and the off-diagonal entries of $\Delta$ are referred to as \textit{within-cluster similarities} and \textit{cross-cluster similarities}, respectively. To elucidate an appropriate choice of the tuning parameter $\lambda$ in the SDP \eqref{eq9}, we adopt the convention that 
\begin{equation*}
    \begin{split}
        p_{n}^{-} &:= \min \left\{ \Delta_{aa} : 1 \leq a \leq k \right\} \\
        &= \binom{s_{\min}-2}{d-2} \left( p_n - q_n \right) + \binom{n-2}{d-2} q_n; \\
        q_{n}^{+} &:= \max \left\{ \Delta_{ab} : 1 \leq a < b \leq k \right\} = \binom{n-2}{d-2} q_n,
\end{split}
\end{equation*}
to denote the \textit{minimum within-cluster similarity} and the \textit{maximum cross-cluster similarity}, respectively. \\
\indent Now, we provide an explicit condition on model parameters $\left(n, p_n, q_n, s_{\min} \right)$ as well as the tuning parameter $\lambda$, under which the solution $\hat{\mathbf{X}}_{\textnormal{SDP}}$ to the convex program \eqref{eq9} is capable of recovering the ground-truth cluster matrix $\mathbf{X}^*$ perfectly. 

\begin{thm} [Performance Guarantee under the Assortative $d$\textsf{-WHSBM}] 
\label{thm4.1} 
Let $\mathbf{A}$ be the similarity matrix of $\mathcal{H} = ([n], \mathbf{W})$ sampled by $d\textnormal{\textsf{-WHSBM}}(n, k, p_n, q_n, \Phi^*)$ with $p_n > q_n$. Then, there is a constant $c_1 > 0$ such that the ground-truth cluster matrix $\mathbf{X}^*$ is the unique optimal solution to the SDP \eqref{eq9} with probability greater than $1-6n^{-11}$, provided that the tuning parameter $\lambda$ satisfies the inequality
\beq
    \label{eq13}
    \frac{1}{4} p_{n}^{-} + \frac{3}{4} q_{n}^{+} \leq \lambda \leq \frac{3}{4} p_{n}^{-} + \frac{1}{4} q_{n}^{+},
\eeq
and model parameters satisfy
\beq
    \label{eq14}
    \begin{split}
    &\ s_{\min}^2 \binom{s_{\min}-2}{d-2}^2 \left( p_n - q_n \right)^2\\
    \geq& \ c_1 \binom{n-2}{d-2} p_n \left( s_{\min} \log n + n \right).
    \end{split}
\eeq
\end{thm}

\noindent Note that \eqref{eq14} has no explicit dependencies on the number of communities $k$ but only through $s_{\min}$.
The proof of Theorem \ref{thm4.1} will be elaborated in Section \ref{proof_thm4.1}. 

\begin{rmk} [Condition \eqref{eq13} on $\lambda$]
\label{remark_on_lambda}
\textnormal{
\hspace{0.1in} To understand the condition on the choice of the tuning parameter $\lambda$ as in \eqref{eq13}, we need to take a closer look at a specific part in the proof of Theorem \ref{thm4.1}: a lower bound on (Q3) in~\eqref{eq25}. One can make an observation that it suffices to take $\lambda$ such that $\left( p_{n}^{-} - \lambda \right) \left( 1 - \mathbf{X}_{ij} \right) \geq c \left( p_{n}^{-} - q_{n}^{+} \right) \left| \mathbf{X}_{ij}^* - \mathbf{X}_{ij} \right|$ for $\mathbf{X}_{ij}^* = 1$, \emph{i.e.}, for pairs of nodes $(i, j)$ belonging to the same cluster, and $\left( \lambda - q_{n}^{+} \right) \mathbf{X}_{ij} \geq c \left( p_{n}^{-} - q_{n}^{+} \right) \left| \mathbf{X}_{ij}^* - \mathbf{X}_{ij} \right|$ for  $\mathbf{X}_{ij}^* = 0$, \emph{i.e.}, for pairs of nodes $(i,j)$ belonging to different communities, for some constant $c > 0$ from \eqref{eq25}. Therefore, we need to choose $\lambda$ such that $\left( p_{n}^{-} - \lambda \right) \wedge \left( \lambda - q_{n}^{+} \right) \geq c \left( p_{n}^{-} - q_{n}^{+} \right)$. In other words, $\lambda$ should lie between the minimum within-cluster similarity $p_n^-$ and the maximum cross-cluster similarity $q_n^+$ to balance the false positives and the false negatives in the variable $\mathbf{X}$. Even though we made a particular choice $c = \frac{1}{4}$ herein which yields the condition \eqref{eq13}, the constant $c$ can be any number in $\left( 0, \frac{1}{2} \right]$ as the absolute constant $c_1$ in Theorem \ref{thm4.1} can be manipulated to be sufficiently large. In conclusion, the condition \eqref{eq13} can be replaced by the weaker one such as $\lambda \in \left[ c p_{n}^{-} + (1-c) q_{n}^{+}, (1-c) p_{n}^{-} + c q_{n}^{+} \right]$ for any $0 < c \leq \frac{1}{2}$.
}
\end{rmk}

\subsection{Comparison with Literature}
\label{comparison_literature}

We next discuss interesting remarks implied by Theorem \ref{thm4.1} and also provide comparisons with existing results.

\begin{rmk}[Sparsity] 
\normalfont{
\label{rmk_sparsity}
We consider the case (a) $p_n = p \alpha_n$ and $q_n = q \alpha_n$ for some constants $p > q > 0$, where the factor $\alpha_n$ stands for \textit{sparsity level} of edge weights, which may depend on $n$. We can deduce that \textsf{CRTMLE} is strongly consistent if $\alpha_n = \Omega \left( \left( n^{d-2} \left( s_{\min} \log n + n \right) \right) / s_{\min}^{2d-2} \right)$ from Theorem \ref{thm4.1}. Now, we impose two additional assumptions on parameters: (b) the number of communities $k$ is constant of $n$; (c) the ground-truth communities are \textit{approximately balanced}, \textit{i.e.}, $s_{\max} / s_{\min} = O(1)$. We remark that this case is studied in \cite{ALS2018}, and we have $s_{\min} = \Theta(n)$ and thus it follows that \textsf{CRTMLE} is strongly consistent if $\alpha_n = \Omega \left( \log n / n^{d-1} \right)$. It coincides with the strong consistency guarantee of \emph{Hypergraph Spectral Clustering with Local Refinement} (\textsf{HSCLR}) \cite{ALS2018}.
}
\end{rmk}

\begin{rmk}[Number of communities] \normalfont{
\label{rmk_numofcomm} 
Suppose that model parameters satisfy the assumption (a) from Remark \ref{rmk_sparsity} with sparsity level $\alpha_n = 1$ (this regime is known as the \textit{dense regime}). We also assume that (b) the communities are \textit{equal-sized}, \textit{i.e.}, $s_{\min} = s_{\max}$. The $d$\textsf{-WHSBM} with parameters obeying (b) is called the \textit{balanced} $d$\textsf{-WHSBM} and denoted by $d\textsf{-WHSBM}^{\textsf{bal}}(n, k, p_n, q_n, \Phi^*)$. Different from Remark \ref{rmk_sparsity}, assume that the number of communities $k$ may scale in $n$. Let $s = n / k$ be the size of each cluster. Then, Theorem \ref{thm4.1} implies that \textsf{CRTMLE} is strongly consistent when $s^{2d-2} = \Omega \left( n^{d-2} \left( s \log n + n \right) \right)$ for this case. From this result, it is easy to see that \textsf{CRTMLE} exactly recovers the hidden partition if $s = \Omega(\sqrt{n})$ (equivalently, $k = \frac{n}{s} = O(\sqrt{n})$), and we find that this result agrees with the performance of the spectral method proposed in \cite{CZ2020}. We emphasize that Thm \ref{thm4.1} is applicable to the weighted case, while the performance analysis in \cite{CZ2020} only considers the \textit{binary-edge case} of $d\textsf{-WHSBM}$.}

\end{rmk}

\begin{rmk}[Order-wise unbalanced community sizes] 
\normalfont{
\label{rmk_unbalancedness} 
Most strong consistency results under variants of hypergraph \textsf{SBM} have been limited to the case in which community sizes are approximately balanced. To the best of our knowledge, this is the first study on the strong consistency for the hypergraph \textsf{SBM} without any such assumptions on cluster sizes. In particular, if the assumption (a) from Remark \ref{rmk_sparsity} is assumed with $\alpha_n = 1$, \textsf{CRTMLE} achieves the strong consistency when $s_{\min} = \Omega(\sqrt{n})$ regardless of $s_{\max}$ by Theorem \ref{thm4.1}. In Section \ref{empirical_results}, we further demonstrate the robustness of \textsf{CRTMLE} against the unbalancedness of community sizes empirically, and compare its performance with other spectral methods.  
}
\end{rmk}

\begin{rmk} [Comparison with the best known result for the graph case] 
\textnormal{
\label{comparison_graph_case}
One can see that the key condition \eqref{eq14} reads $\frac{p_n - q_n}{\sqrt{p_n}} \gtrsim \left( \frac{\sqrt{n}}{s_{\min}} \right)^{d-2} \max \left\{ \frac{\sqrt{n}}{s_{\min}}, \sqrt{\frac{\log n}{s_{\min}}} \right\}$. Now, we consider the binary-edge case of the graph \textsf{SBM}. The best known result for the exact recovery is $\frac{p_n - q_n}{\sqrt{p_n}} \gtrsim \max \left\{ \frac{\sqrt{n}}{s_{\min}} \cdot \sqrt{\frac{q_n}{p_n}}, \sqrt{\frac{\log n}{s_{\min}}} \right\}$ \cite{CL2015} in this model. Under the parameter regime $p_n / q_n = \Theta(1)$, which encompasses the most challenging regime of the \textsf{SBM}s, our result matches with the best known result.}
\end{rmk}

\begin{rmk} [Tightness]
\label{rmk_tightness}
\textnormal{
Our proposed algorithm is both computationally feasible and information-theoretically optimal under the assumptions (a), (b) and (c) in Remark \ref{rmk_sparsity}, since it has been proven in \cite{CKK2015} that there is no strongly consistent estimator when $\alpha_n = o \left( \log n / n^{d-1} \right)$ and \textsf{CRTMLE} achieves the strong consistency if $\alpha_n = \Omega \left( \log n / n^{d-1} \right)$. On the other hand, it is still an open problem whether $s_{\min} = \Omega(\sqrt{n})$ gives the order-wise computational limit in the setting assumed in Remark \ref{rmk_unbalancedness} even for the graph case \cite{CL2015, CX2016, CSX2014, OH2011, A2014}. We argue in Section \ref{effect_outliers} that it is possible to exactly recover the hidden clique of size $s = \Omega (\sqrt{n})$ within poly-time in the planted clique model for hypergraphs. Even though it has been conjectured that $s = o(\sqrt{n})$ is a computationally-hard regime \cite{AAKMRX2007, R2010, FGRVX2017}, proving this conjecture rigorously still remains open. Furthermore, no general thresholds, \emph{i.e.}, converse result of Theorem \ref{thm4.1} or the information-theoretic limit, are known for general parameter regimes of  $(n, k, p_n, q_n, \Phi^*)$.
}
\end{rmk}

\subsection{Estimating the Tuning Parameter in the Balanced Case} 
\label{estimation_lambda}

\begin{algorithm}[b]
	\caption{Estimation of $\lambda$ from the Observed Data}
	\label{algorithm2}

\begin{algorithmic}[1]
\STATE \textbf{Data}: The observed similarity matrix $\mathbf{A} \in \mathbb{R}^{n \times n}$.
\STATE Compute and sort the eigenvalues of $\mathbf{A}$; denote them by $\hat{\lambda}_{i} := \lambda_{i}(\mathbf{A})$ for each $i \in [n]$.
\STATE Let $\hat{k} := \arg \max \left\{ \hat{\lambda}_{i} - \hat{\lambda}_{i+1} : i \in \left\{ 2, 3, \cdots, n-1 \right\} \right\}$ (broken tie uniformly at random).
\STATE Set $\hat{s} := \frac{n}{\hat{k}}$, $\hat{p}_{n}^{-} := \frac{\hat{s} \hat{\lambda}_{1} + (n - \hat{s}) \hat{\lambda}_{2}}{n \left( \hat{s} - 1 \right)}$ and $\hat{q}_{n}^{+} := \frac{\hat{\lambda}_{1} - \hat{\lambda}_{2}}{n}$.
\STATE \textbf{Output}: An estimator $\hat{\lambda} := \frac{\hat{p}_{n}^{-} + \hat{q}_{n}^{+}}{2}$ of $\lambda$.
\end{algorithmic}
\end{algorithm}

\indent Our algorithm \textsf{CRTMLE} requires an extraneous input $\lambda$. For its success, we need to make a suitable choice of the tuning parameter $\lambda$ so that it obeys the bound \eqref{eq13}. In this section, we consider the \textit{balanced case} of the $d$\textsf{-WHSBM} ($s_{\min} = s_{\max}$) and provide an algorithm (Algorithm \ref{algorithm2}) to specify the tuning parameter $\lambda$ in a completely data-driven way by estimating the model parameters $(k, p_n, q_n)$ with strong theoretical guarantees. Let  $d\textsf{-WHSBM}^{\textsf{bal}}(n, k, p_n, q_n, \Phi^*)$ denote the balanced model and $s$ denote the size of communities. \\
\indent Algorithm \ref{algorithm2} is built upon the observation that the eigenvalues of $\mathbb{E}[\mathbf{A}]$ are given by
\beq
    \label{eq15}
    \begin{split}
    &\ \lambda_{i} := \lambda_{i}(\mathbb{E}[\mathbf{A}]) 
    \\
    =& \ 
    \begin{cases}
        (s-1) \left( p_{n}^{-} - q_{n}^{+} \right) + (n-1) q_{n}^{+} & \textnormal{when } i = 1; \\
        (s-1) \left( p_{n}^{-} - q_{n}^{+} \right) - q_{n}^{+} & \textnormal{when } 2 \leq i \leq k; \\
        - p_{n}^{-} & \textnormal{when } k+1 \leq i \leq n.
    \end{cases}
    \end{split}
\eeq
\noindent Similar idea is utilized to setup the tuning parameter of SDP for graph clustering in the \textsf{SBM} \cite{CSX2014}. \\
\indent Theorem below guarantees that the errors of the estimators $\hat{k}$, $\hat{s}$, $\hat{p}_{n}^{-}$ and $\hat{q}_{n}^{+}$ from Algorithm \ref{algorithm2} are sufficiently small and the estimator $\hat{\lambda}$ of $\lambda$ satisfies the desired condition \eqref{eq13} in Theorem \ref{thm4.1}. Now we state the formal result below, deferring the proof to Appendix \ref{appendix:a}.

\begin{thm} [Accuracy of Estimators from Algorithm \ref{algorithm2}] 
\label{thm4.2}
Let $\mathbf{A}$ denote the similarity matrix of $\mathcal{H} = ([n], \mathbf{W})$ generated by $d\textnormal{\textsf{-WHSBM}}^{\textnormal{\textsf{bal}}}(n, k, p_n, q_n, \Phi^*)$. Suppose that the condition \eqref{eq14} holds with a sufficiently large constant $c_2 > 0$. Then, the estimators computed in Algorithm \ref{algorithm2} satisfy the following properties with probability exceeding $1 - 4n^{-11}$\textnormal{:}
\beq \nonumber
\begin{split}
&\text{1. } \hat{k} = k \text{ and } \hat{s} = s, \\
&\text{2. } \max \left\{ \left| \hat{p}_{n}^{-} - p_{n}^{-} \right|, \left| \hat{q}_{n}^{+} - q_{n}^{+} \right| \right\} \leq \frac{2 c_5}{s} \sqrt{n \binom{n-2}{d-2} p_n},\\
&\text{3. } \hat{\lambda} \in \left[ \frac{1}{4} p_{n}^{-} + \frac{3}{4} q_{n}^{+}, \frac{3}{4} p_{n}^{-} + \frac{1}{4} q_{n}^{+} \right].
\end{split}
\eeq
Here, the constant $c_5$ in Cor. \ref{cor7.1} is specified with $\alpha = 1$.
\end{thm}

\indent Merging Algorithm \ref{algorithm1} and \ref{algorithm2}, we get a complete polynomial-time algorithm, which identifies the hidden partition of $[n]$ \textit{w.h.p.} in the balanced $d$\textsf{-WHSBM} without any prior knowledge of model parameters $\left( k, p_n, q_n \right)$.

\section{Model Extensions}
\label{model_extensions}

\indent In this section, we study two important variations of the $d$\textsf{-WHSBM} to reflect circumstances where edge weights are partially observed or outlier nodes may exist, and we analyze our algorithm under these modified models.

\subsection{Clustering Partially Observed Weighted Hypergraphs}
\label{partially_observed}
\begin{table*}[h]
    \centering    
    \caption{Comparison with existing results on exact recovery under the \textsf{SBM} with partial observations.}
    \label{tab:table2}
    \begin{tabular}{|c|c|c|}
        \hline
        Paper & Condition on $\left( p_n, q_n, \varepsilon_{n}, s_{\min} \right)$ & Algorithm \\
        \hline \hline 
        \cite{OH2011} & $\left( p_n - q_n \right) \varepsilon_{n} \gtrsim \frac{\sqrt{n}}{s_{\min}}$ & ``Low-rank $+$ Sparse'' Decomposition \\ [0.3ex]
        \hline
        \cite{CJSX2014} & $\left( 1 - 2 \tau_{n} \right) \sqrt{\varepsilon_{n}} \gtrsim \frac{\sqrt{n} \log n}{s_{\min}}$ & ``Low-rank $+$ Sparse'' Decomposition \\ [0.3ex]
        \hline
        \cite{CSX2014} & $\left( p_n - q_n \right) \sqrt{\frac{\varepsilon_n}{p_n}} \gtrsim \max \left\{ \frac{\sqrt{n}}{s_{\min}}, \frac{(\log n)^2}{\sqrt{s_{\min}}} \right\}$ & Convexified MLE \\ [0.3ex]
        \hline \hline
        This paper & \multirow{2}{*}{$\left( p_n - q_n \right) \sqrt{\frac{\varepsilon_n}{p_n}} \gtrsim \max \left\{ \frac{\sqrt{n}}{s_{\min}}, \sqrt{\frac{\log n}{s_{\min}}} \right\}$} & \multirow{2}{*}{Convexified MLE} \\
        (Corollary \ref{cor5.1}) & & \\
        \hline
    \end{tabular}
\end{table*}
\indent We consider the case where multi-way relations among the nodes are \textit{partially observed}. A standard and extensively used model for clustering partially-observed unweighted graphs is a random graph model with missing data, also known as the \textsf{SBM} \emph{with partial observations} \cite{OH2011, CJSX2014, CSX2014, VOH2014}. We extend this model to the weighted hypergraph case as follows: First, consider a weighted hypergraph $\mathcal{H} = \left( [n], \mathbf{W} \right)$ sampled by $d\textsf{-WHSBM}(n, k, p_n, q_n, \Phi^*)$. Each entry of $\mathbf{W}$ is observed independently with probability $\varepsilon_{n}$. We let $\mathbf{W}^{\textnormal{obs}} := \left( W_{e}^{\textnormal{obs}} : e \in \mathcal{E} \right)$ denote the observed weighted $d$-uniform hypergraph, \textit{i.e.}, for each $e \in \mathcal{E}$, the associated weight $W_{e}^{\textnormal{obs}}$ is given by $W_{e}^{\textnormal{obs}} = W_{e}\in[0,1]$ if the entry $W_e$ is observed, and $W_{e}^{\textnormal{obs}} = \times$ otherwise. We refer to this model as the $d$\textsf{-WHSBM} with partial observations with parameters $n, k \in \mathbb{N}$, $0 \leq q_n < p_n \leq 1$, $\varepsilon_{n} \in [0, 1]$ and $\Phi^* : [n] \rightarrow [k]$. \\
\indent Our main goal is to recover the latent membership structure with partial observations. This clustering problem with missing data can be solved efficiently via the following two-stage procedure. First, set to zero all the unobserved entries of $\mathbf{W}^{\textnormal{obs}}$ and let $\mathbf{W}'$ denote the weighted $d$-uniform hypergraph obtained by zeroing-out the unobserved entries of $\mathbf{W}^{\textnormal{obs}}$. Then, we perform \textsf{CRTMLE} on $\mathcal{H}' := \left( [n], \mathbf{W}' \right)$. The \textit{zero-imputed weighted $d$-uniform hypergraph} $\mathcal{H}' = \left( [n], \mathbf{W}' \right)$ of $\mathcal{H}^{\textnormal{obs}}$ can be viewed as a data generated by the $d\textsf{-WHSBM}(n, k, p_n \varepsilon_{n}, q_n \varepsilon_{n}, \Phi^*)$, so that the theoretical guarantee of this two-stage method is obtained from Theorem \ref{thm4.1} immediately.

\begin{cor} [Performance Guarantee under the $d$\textsf{-WHSBM} with Missing Data] 
\label{cor5.1}
Let $\mathcal{H}' = \left( [n], \mathbf{W}' \right)$ be the zero-imputed data of a sample $\mathcal{H}^{\textnormal{obs}} = \left( [n], \mathbf{W}^{\textnormal{obs}} \right)$ drawn from the $d$\textnormal{\textsf{-WHSBM}} with partial observations and $\mathbf{A}'$ denote its similarity matrix. Then, the SDP \eqref{eq9} applied to $\mathbf{A}'$ with a tuning parameter $\lambda$ obeying
\begin{equation*}
    \frac{1}{4} p_{n}^{-} \varepsilon_{n} + \frac{3}{4} q_{n}^{+} \varepsilon_{n} \leq \lambda \leq \frac{3}{4} p_{n}^{-} \varepsilon_{n} + \frac{1}{4} q_{n}^{+} \varepsilon_{n},
\end{equation*}
recovers the ground-truth cluster matrix $\mathbf{X}^*$ exactly with probability at least $1 - 6n^{-11}$, when
\begin{equation*}
    \left( p_n - q_n \right) \sqrt{\frac{\varepsilon_{n}}{p_n}} \geq c_2 \left( \frac{\sqrt{n}}{s_{\min}} \right)^{d-2} \max \left\{ \frac{\sqrt{n}}{s_{\min}}, \sqrt{\frac{\log n}{s_{\min}}} \right\},
\end{equation*}
where $c_2 > 0$ is an absolute constant. Here, $p_{n}^{-}$ and $q_{n}^{+}$ refer to the minimum within-cluster similarity and the maximum cross-cluster similarity, respectively. (See Section \ref{performance_analysis} for details.)
\end{cor}

\indent To the best of our knowledge, there has been no provable computationally-feasible method for clustering partially observed hypergraphs for general parameters. Nonetheless, there are some eminent works on clustering graphs with missing data. Their common framework is the graph \textsf{SBM} with partial observations. Since Cor. \ref{cor5.1} is applicable to this model, we may compare it with previous works. See Table \ref{tab:table2} for a summary of comparison with literature. In this table, we let $\tau_n := \max \left\{ 1 - p_n, q_n \right\}$ and thus $1 - 2\tau_n$ is a lower bound of the density gap $p_n - q_n$. One can see that our analytical result is either as good as or order-wise stronger than existing works when $s_{\min} \lesssim n / \log n$. If $s_{\min} \gtrsim n / \log n$, our result is order-wise better than ones in \cite{CJSX2014, CSX2014}.

\subsection{Effect of Outlier Nodes}
\label{effect_outliers}

\indent In this subsection, we show that \textsf{CRTMLE} is robust against the presence of outlier nodes in the balanced $d$\textsf{-WHSBM}.

\subsubsection{Robustness Against Outlier Nodes in the Balanced Case}

The robustness of \textsf{CRTMLE} against the unbalancedness of cluster sizes is a crucial benefit, compared to spectral methods, as summarized in Remark \ref{rmk_unbalancedness}. Another strength of convex relaxation methods is the robustness against outlier nodes \cite{LCX2018}, which will be elaborated in this subsection for the balanced case. There are several existing works studying the effect of outliers \cite{CX2016, CSX2014, CL2015} for graph clustering, but not many for hypergraph clustering. \\
\indent First, consider a new framework for hypergraph clustering that allows the presence of \textit{outlier nodes}. Let $\mathcal{V} := [n] = \mathcal{I} \cup \mathcal{O}$ be the set of $n$ nodes, where $\mathcal{I}$ is the set of inlier nodes, while $\mathcal{O}$ denotes the set of outlier nodes. These nodes are endowed with the following latent  membership structure: each inlier node $i \in \mathcal{I}$ is labeled with community assignment $\Phi^*(i) \in [k]$, while every outlier node $o \in \mathcal{O}$ is simply labeled by $\Phi^*(o) = k+1$. We denote $k$ hidden clusters and their sizes by $C_{a}^* := \left( \Phi^* \right)^{-1}(a)$ and $s_{a} := \left| C_{a}^* \right|$, $a \in [k]$, respectively. Also, we use the convention that $C_{k+1}^* := \left( \Phi^* \right)^{-1}(k+1) = \mathcal{O}$ and $s_{k+1} := \left| C_{k+1}^* \right| = n - \sum_{a=1}^{k} s_a$, and emphasize that $C_{k+1}^*$ is not indeed an underlying community. Let $s_{\min} := \min \left\{ s_a : a \in [k] \right\}$ and $s_{\max} := \max \left\{ s_a : a \in [k] \right\}$. We assume that all communities are equal-sized, \textit{i.e.}, $s = s_{\min} = s_{\max}$. \\
\indent We first extend the $\Phi^*$-homogeneity of each $d$-regular edge $e \in \mathcal{E}$ and say that a $d$-regular edge $e \in \mathcal{E}$ is \textit{$\Phi^*$-homogeneous} if $e \subseteq C_{a}^*$ for some $a \in [k]$, and \textit{$\Phi^*$-heterogeneous} otherwise. We describe our main framework involving five model parameters, $n, k \in \mathbb{N}$, $0 \leq q_n < p_n \leq 1$ and $\Phi^* : [n] \rightarrow [k+1]$. A weighted $d$-uniform hypergraph $\mathcal{H} = \left([n], \left( W_{e} : e \in \mathcal{E} \right) \right)$ is generated as follows: $W_e \in [0, 1]$ is assigned to each $e \in \mathcal{E}$ independently such that $\mathbb{E}[W_e] = p_n$ if $e$ is $\Phi^*$-homogeneous, and $\mathbb{E}[W_e] = q_n$ otherwise. We call this model the \textit{balanced} $d$\textsf{-WHSBM} \textit{with outlier nodes}. It can also be referred to as the \textit{weighted $d$-uniform hypergraph planted clustering model} ($d$\textsf{-WHPCM}) by adopting terminologies from \cite{CX2016}. \\
\indent For subsequent discussion, we modify a matrix representation of the latent membership structure from Section \ref{algorithm_description}, which reflects the presence of outlier nodes. Define the \textit{ground-truth membership matrix} $\mathbf{Z}^* \in \left\{ 0, 1 \right\}^{n \times k}$ defined by $\mathbf{Z}_{ia}^* = 1$ if $\Phi^*(i) = a$ and $\mathbf{Z}_{ia}^* = 0$ otherwise. Note that $\mathbf{Z}_{i*} = 0$ if and only if $i \in [n]$ is an outlier node. Then, we can represent the membership structure by the \textit{ground-truth cluster matrix} $\mathbf{X}^* := \mathbf{Z}^* \left( \mathbf{Z}^* \right)^{\top}$, where $\mathbf{X}_{ii}^* = 1$ for $i \in \mathcal{I}$, $\mathbf{X}_{ii}^* = 0$ for $i \in \mathcal{O}$, and for every $i \neq j$, $\mathbf{X}_{ij}^* = 1$ if and only if $i$ and $j$ belong to the same community. \\
\indent We now assert that the SDP \eqref{eq10} is robust against the outlier nodes under the balanced case. In this case, the constraints of \eqref{eq10} become $\langle \mathbf{I}_{n}, \mathbf{X}^* \rangle = ks$ and $\langle \mathbf{1}_{n \times n}, \mathbf{X}^* \rangle = ks^2$. So, with the exact knowledge of the number of communities $k$ and their size $s$, we can implement the SDP \eqref{eq10}. With this alternative SDP, we obtain a provable polynomial-time algorithm which identifies the hidden communities from an observed data generated by the $d$\textsf{-WHPCM}. The proof of the following result closely follows the proof of Thm \ref{thm4.1}, so we omit the details. 

\begin{thm} [Performance Guarantee in the $d$\textsf{-WHPCM}] 
\label{thm5.1}
Let $\mathbf{A}$ be the similarity matrix of a sample $\mathcal{H}$ drawn from $d\textnormal{\textsf{-WHPCM}}(n, k, p_n, q_n, \Phi^*)$. Then, there is a universal constant $c_4 > 0$ (replacing $c_1 > 0$ in \eqref{eq14}) such that the ground-truth cluster matrix $\mathbf{X}^*$ is the unique optimal solution to the SDP \eqref{eq10} with prob. greater than $1 - 6n^{-11}$, when parameters obey the condition \eqref{eq14}. 
\end{thm}

\subsubsection{Planted Clique Problem for Hypergraphs}

One prominent planted problem is the \textit{standard planted clique problem} \cite{AKS1998}. We now generalize this problem for hypergraphs: the task of finding a hidden clique of size $s$, that has been planted in an Erd\"{o}s-R\'{e}nyi (\textsf{ER}) model for random $d$-uniform hypergraphs.

\begin{defi} [The Planted Clique Model for Hypergraphs] 
\label{def5.1}
\normalfont{
Let $\mathcal{C} \subseteq [n]$ a hidden subset of size $s \leq n$. A $d$-uniform hypergraph $\mathcal{H} = \left( [n], E(\mathcal{H}) \right)$, where $E(\mathcal{H}) \subseteq \mathcal{E}$, is generated as follows: each $d$-regular edge $e \in \mathcal{E}$ appears independently as an hyperedge of $\mathcal{H}$ with prob. 1 if $e \subseteq \mathcal{C}$, and $\frac{1}{2}$ otherwise.
}
\end{defi}
\indent Observe that the binary-edge case of the $d$\textsf{-WHPCM}, with $k = 1$, $p_n = 1$ and $q_n = \frac{1}{2}$, retrieves the above model. Hence, Theorem \ref{thm5.1} ensures the exact recovery of the hidden clique for the size $s = \Omega(\sqrt{n})$ regardless of the value of $d$, and we remark that this result is consistent with the state-of-the-art bound for the graph case \cite{AKS1998, CSX2014, CX2016, CL2015}.

\section{Empirical Results}
\label{empirical_results}

\indent In this section, we provide simulation results demonstrating the robustness of our proposed algorithm against the unbalancedness of community sizes (Section \ref{robust_unbalancedness}) as well as the presence of outliers (Section \ref{robust_outliers}). We also conduct a set of simulations to show the performance and robustness of the proposed algorithm for a real application of hypergraph clustering in computer vision, the subspace clustering \cite{ALZPKB2005} in Section~\ref{subspace_clustering}. \\
\indent From all these experimental results, we are able to confirm that \textsf{CRTMLE} outperforms the state-of-the-arts for hypergraph clustering, especially as community sizes become more unbalanced or the number of outliers increases. \\
\indent We compare the performance of our algorithm with several state-of-the-art algorithms including \textsf{TTM} \cite{GD2015-1}, \textsf{NH-Cut} \cite{ZHS2007}, \textsf{HOSVD} \cite{G2005}, \textsf{HSCLR} \cite{ALS2018}, and \textsf{hMETIS} \cite{KK2000}. All of these algorithms hinge upon either spectral property of the similarity matrix or the graph partitioning method, whereas our algorithm is based on the SDP relaxation. 

\subsection{Robustness Against the Unbalancedness of Community Sizes}
\label{robust_unbalancedness}

\indent Let us fix the size of hyperedges as $d=3$ and use Bernoulli distribution with mean $p_n$ for generating homogeneous hyperedges, and $q_n$ for heterogeneous hyperedges, where
\begin{equation*}
    p_n =  p \cdot \frac{n \log n}{\binom{n}{d}} \quad \textrm{and} \quad q_n = q \cdot \frac{n \log n}{\binom{n}{d}},
\end{equation*}
with some constants $p > q>0$ which we will specify. We set $(n, p) \in \{144,288,432,576\} \times \{10,15,20,25\}$, where $q$ is fixed to $5$, and set the number of clusters to be $k \in \left\{ 3, 4 \right\}$.
\begin{enumerate}
    \item $k = 3$: three different combinations of community sizes
    $\left\{ n/3, n/3, n/3 \right\}$, $\left\{ n/6, n/3, n/2 \right\}$, $\left\{ n/12, n/3,  7n/12 \right\}$ are considered to represent the different levels of unbalancedness.
    \item $k = 4$: two different combinations $\left\{ n/4, n/4, n/4, n/4 \right\}$ and $\left\{ n/12, n/6, n/3, 5n/12 \right\}$
    are considered for the balanced and unbalanced community sizes, respectively.
\end{enumerate}
\begin{figure}
    \centering
    \includegraphics[width = \columnwidth]{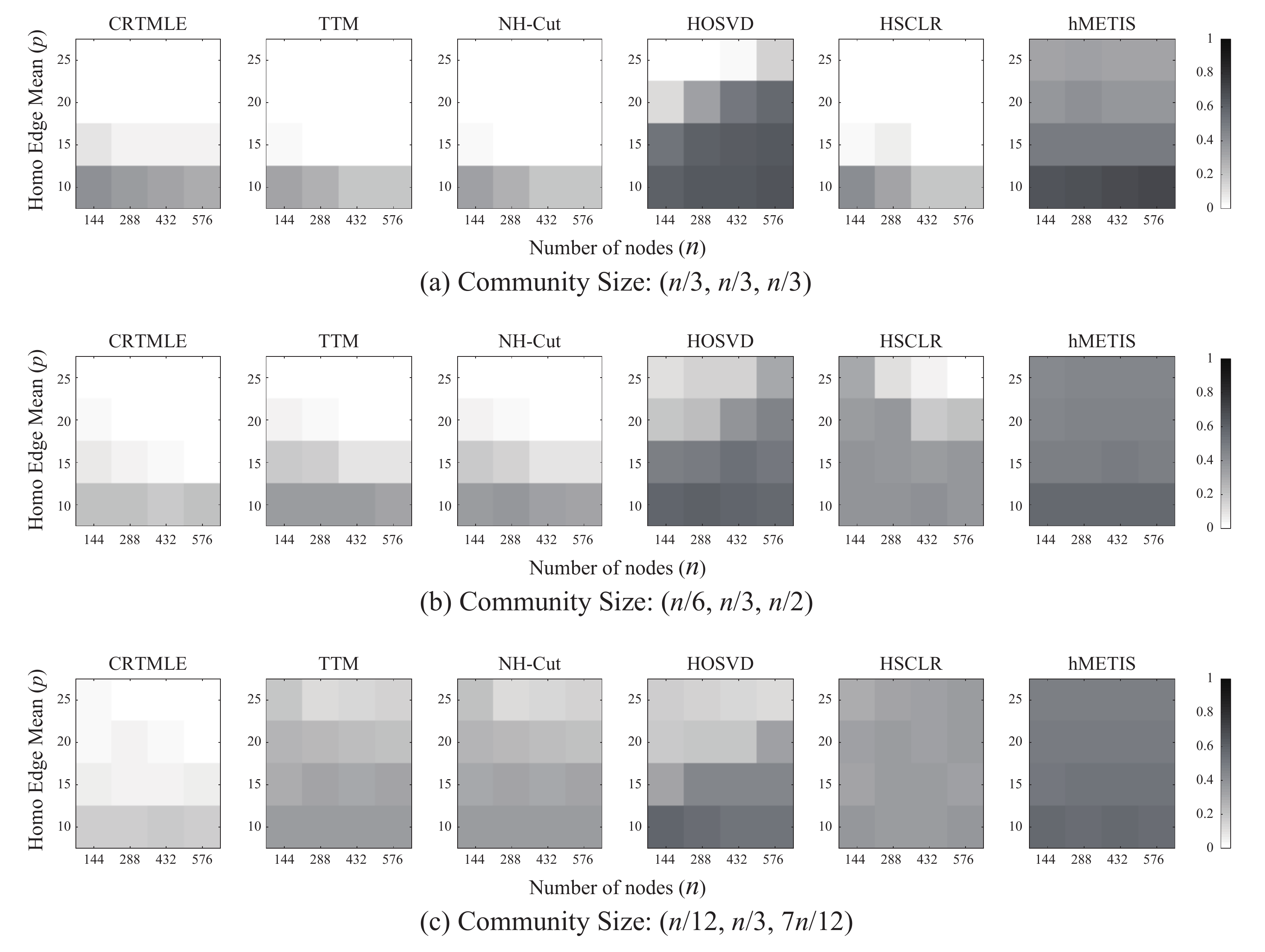}
    \caption{Empirical performance (the average fractional errors over 50 trials) of the proposed algorithm \textsf{CRTMLE} (most left) compared to other state-of-the-art algorithms for hypergraph clustering for the number of communities $k = 3$. The community sizes become more unbalanced from (a) to (c). A lighter color implies a lower fractional error.}
    \label{fig:simulation1}
\end{figure}

\begin{figure}
    \centering
    \includegraphics[width = \columnwidth]{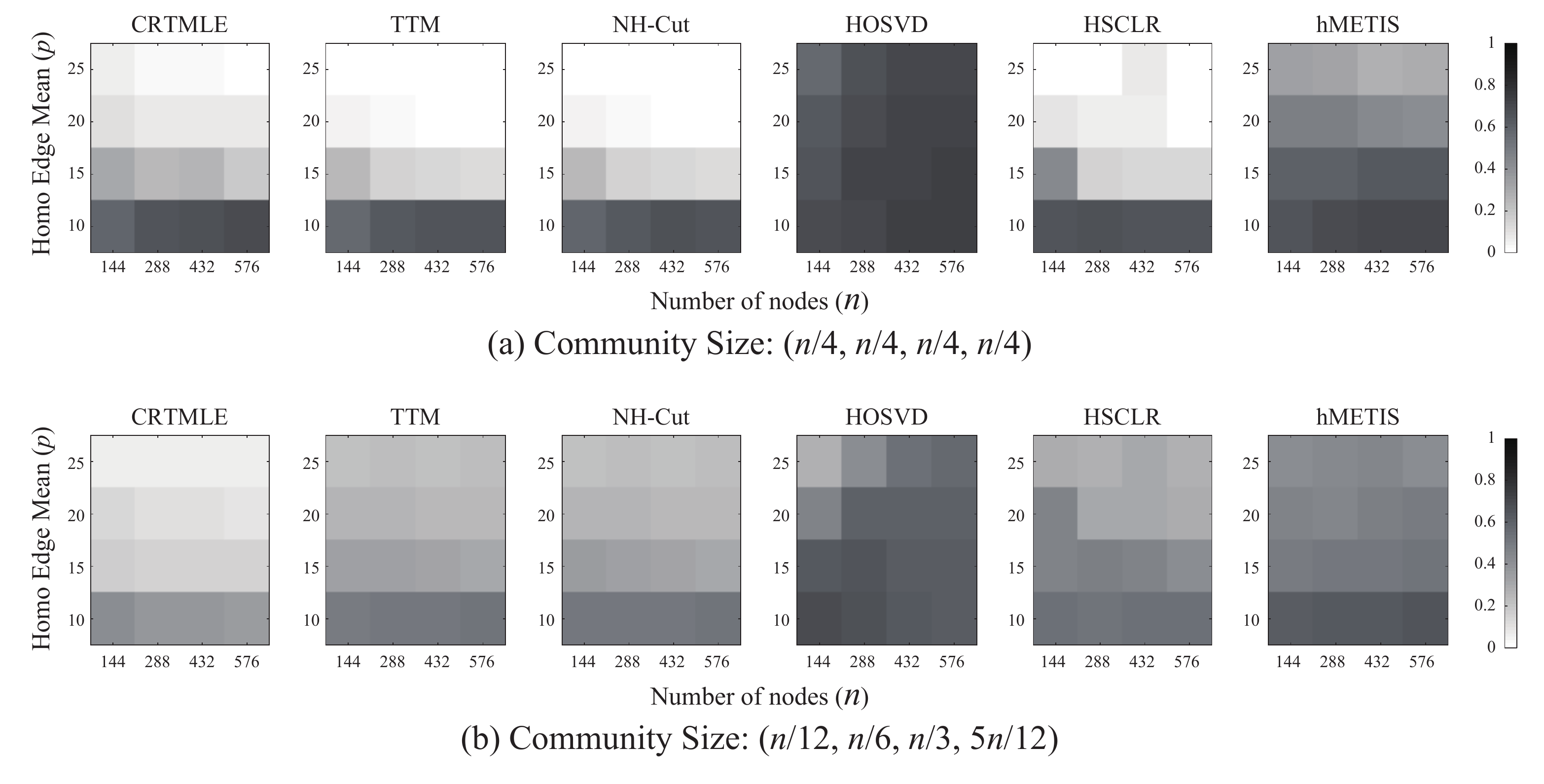}
    \caption{Empirical performance (the average fractional errors over 50 trials) of the proposed algorithm \textsf{CRTMLE} (most left) compared to other state-of-the-art algorithms for hypergraph clustering for the number of communities $k = 4$. The community sizes become more unbalanced from (a) to (c). A lighter color implies a lower fractional error.}
    \label{fig:simulation2}
\end{figure}

\indent We run each algorithm 50 trials on randomly generated hypergraphs and measure the fractional errors of each algorithm. When implementing \textsf{CRTMLE} (Alg. \ref{algorithm1}), we set the tuning parameter $\lambda= \frac{p_{n}^{-} + q_{n}^{+}}{2}$, where $p_{n}^{-}$ and $q_{n}^{+}$ are the minimum within-cluster similarity and the maximum cross-cluster similarity, respectively, as defined in Section \ref{performance_analysis}. The experimental results are summarized in Figure \ref{fig:simulation1} for the case $k = 3$, and in Figure \ref{fig:simulation2} for the case $k = 4$, respectively. In the figures, a lighter color implies a lower fractional error. We can observe that \textsf{CRTMLE} shows comparable performance with other algorithms when the community sizes are balanced, but shows the best performance among all the algorithms in most parameter regimes where the community sizes are unbalanced. Especially, the performances of other algorithms degrade as the community sizes become more unbalanced, while \textsf{CRTMLE} has almost consistent performance regardless of the unbalancedness of the community sizes. This result matches with Remark \ref{rmk_unbalancedness}, where we explain that \textsf{CRTMLE} is robust against the heterogeneity in community sizes.

\subsection{Robustness Against the Presence of Outlier Nodes}
\label{robust_outliers}

\begin{figure}
    \centering
    \includegraphics[width = \columnwidth]{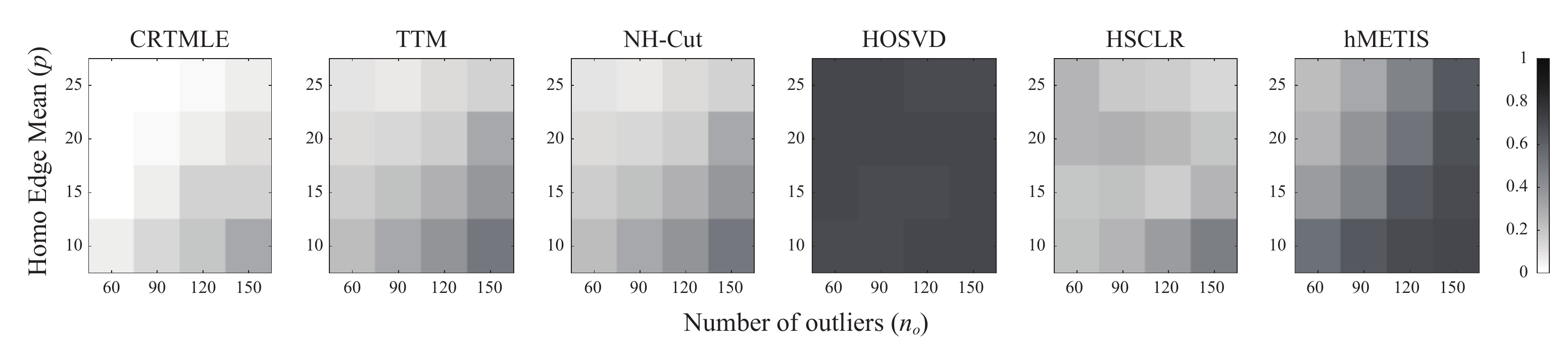}
    \caption{Empirical performance (the average fractional errors over 50 trials) of the proposed algorithm \textsf{CRTMLE} (most left) compared to other state-of-the-art algorithms for hypergraph clustering for the number of communities $k = 3$. A lighter color implies a lower fractional error. As the number of outlier nodes ($n_0$: $x$-axis) increases, the fractional error increases for all the algorithms, but \textsf{CRTMLE} is the most robust algorithm against the outlier nodes.
    }
    \label{fig:simulation3}
\end{figure}

\indent In the next simulation, we add $n_o$ outlier nodes for the case of equal-sized communities with $k = 3$ and $n = 300$. A similar setting with previous ones in Section \ref{robust_unbalancedness} is assumed, except that $\left( n_o, p \right) \in \{ 60, 90, 120, 150 \} \times \{ 10, 15, 20, 25 \}$, and $q = 1$. As expected, all methods excepting \textsf{CRTMLE} degrade as the number of outlier nodes increases, but \textsf{CRTMLE} is the most robust one against the outlier nodes, as shown in Figure \ref{fig:simulation3}.

\subsection{An Application of Hypergraph Clustering: Subspace Clustering}
\label{subspace_clustering}

\begin{figure}
    \centering
    \includegraphics[width = \columnwidth]{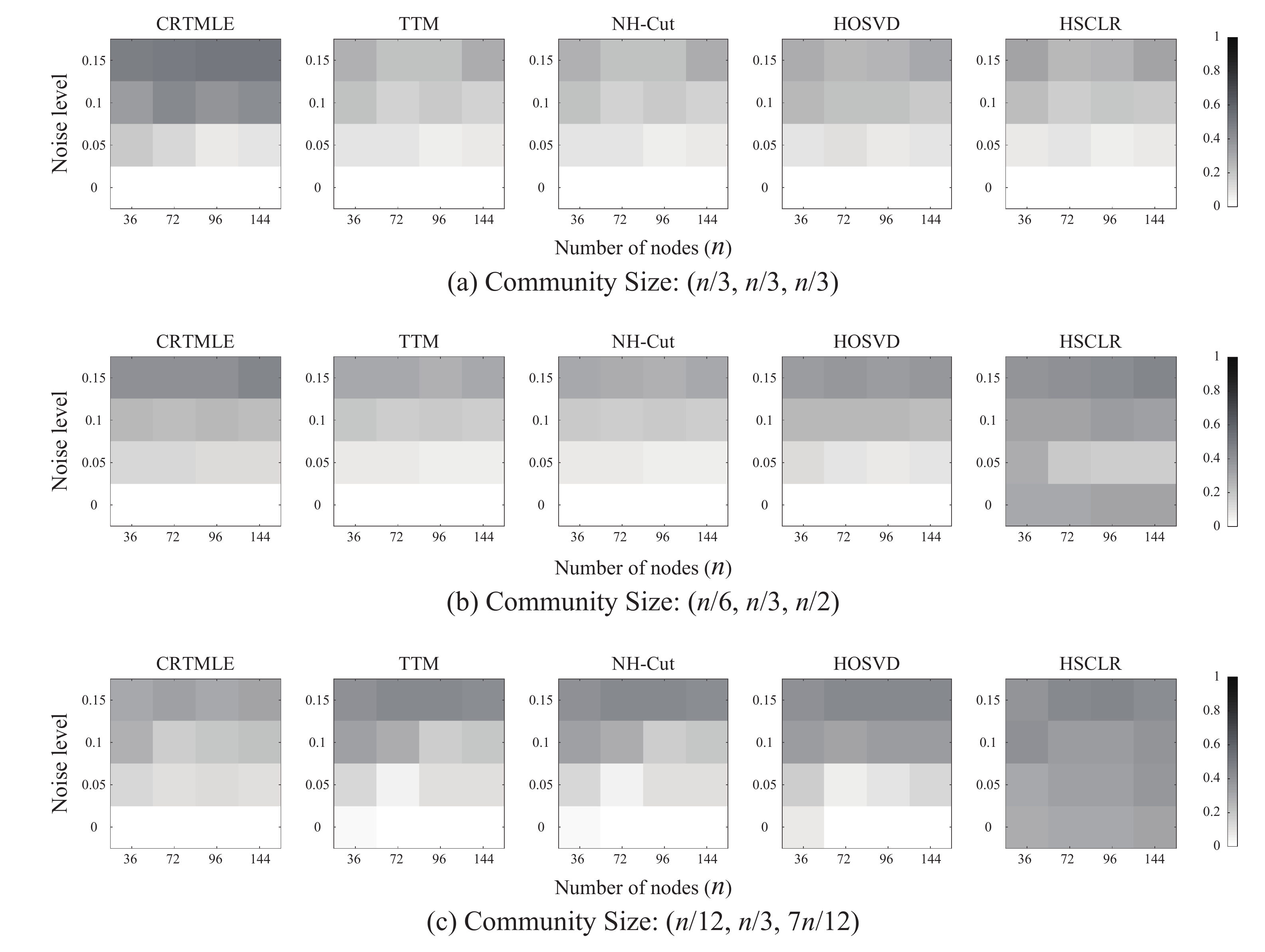}
    \caption{Empirical performance (the average fractional errors over 50 trials) of the proposed algorithm \textsf{CRTMLE} (most left) compared to other state-of-the-art algorithms for subpace clustering for the number of subspaces $k = 3$. The cluster sizes (the number of points in each subspace) become more unbalanced from (a) to (c). A lighter color implies a lower fractional error. The \textsf{CRTMLE} is the most robust algorithm against the unbalancedness of cluster sizes.
    }
    \label{fig:simulation4}
\end{figure}

\indent For the last experiment, we apply the hypergraph clustering to solve the subspace clustering problem \cite{ALZPKB2005}, which has wide applications in computer vision. In subspace clustering, each cluster is formed by points that (closely) lie on the same subspace. The goal is to recover these clusters by measuring some similarities between $d$ data points and applying the hypergraph clustering on the generated weighted hypergraphs. The weights of the hypergraph $W_{e}\in [0,1]$, $e \in \mathcal{E}$, indicate the ``fitness'' of the $d$ points $\left\{ i_1, i_2, \cdots, i_d \right\} $ by a hyperplane and the higher $W_{e}$ implies the better the $d$ points can be fitted by a subspace. 
In this experiment, we use the simulation setup similar to \cite{GD2015-1, GD2017-1}. In an ambient space of dimension $3$, we randomly generate three 1-dimensional subspaces (lines) and sample a fixed number of data points from each subspace. After that, a zero-mean Gaussian noise vector is added to every data point, where the covariance matrix of the noise vector is $\sigma \mathbf{I}_3$. The noise parameter $\sigma$ indicates the level of difficulty in subspace clustering. A uniform hypergraph with $d=3$ is then generated by calculating polar curvature of every three data points, which quantifies how close the three points can be fitted to a line. We then compare the performance of \textsf{CRTMLE} with other four state-of-the-art hypergraph clustering algorithms in recovering $k=3$ clusters for $n\in \{36, 72, 96, 144\}$ and the varying noise level $\sigma$. We run each algorithm for $50$ trials and plot the average of fractional errors for three different combinations of community sizes  $\left\{ n/3, n/3, n/3 \right\}$, $\left\{ n/6, n/3, n/2 \right\}$, and $\left\{ n/12, n/3,  7n/12 \right\}$ in Fig. \ref{fig:simulation4}. The result shows a similar trend with the first simulation: \textsf{CRTMLE} has comparable (or sometimes worse) performance to other algorithms for equal-sized case, but it is robust against the unbalancedness of cluster sizes and thus performs better as the unbalancedness becomes significant.

\section{Proof of Main Result}
\label{proof_mainresult}

\subsection{Concentration Bounds of Spectral Norm}
\label{concentration_bounds}

Before we get into the proof of Theorem \ref{thm4.1}, we derive a sharp concentration bound on the spectral norm $\left\| \mathbf{A} - \mathbb{E}[\mathbf{A}] \right\|$ in a specific parameter regime of the assortative $d$\textsf{-WHSBM}, which will be elaborated below. It plays a crucial role in proving the main results. In spectral method and SDP analysis, it has been a technical challenge to obtain a tight probabilistic bound on the spectral norm of random matrices. Since standard random matrix theory used in concentration results of the adjacency matrix of graph \textsf{SBM} mainly assumes the independence between entries, they cannot be directly employed for the similarity matrix $\mathbf{A}$ under the $d$\textsf{-WHSBM}, which has \emph{strong dependencies} across entries due to its construction. In \cite{KBG2018}, the authors employ the \emph{matrix Bernstein inequality} \cite{T2012} to resolve such a dependency issue occured in a variation of the graph Laplacian matrix. Unfortunately, this approach is not strong enough to settle the desired tight bound on $\mathbf{A}$ for a wider range of parameter regimes considered in this paper. To be specific, utilizing the matrix Bernstein inequality on the decomposition $\mathbf{A} = \sum_{e \in \mathcal{E}} W_{e} \mathbf{M}_{e}$, where $\mathbf{M}_{e} := \sum_{i, j \in e : i \neq j} \mathbf{e}_{i} \mathbf{e}_{j}^{\top}$ for $e \in \mathcal{E}$, gives $\| \mathbf{A} - \mathbb{E} [\mathbf{A}] \| =  O \left( \sqrt{n \binom{n-2}{d-2} p_n \cdot \log n} \right)$. This results in a weaker (sub-optimal) result than what we will demonstrate to prove Theorem \ref{thm4.1}. \\
\indent One of our technical contribution is in providing a tighter spectral norm bound on $\mathbf{A}$ via the Friedman-Kahn-Szemer\'{e}di argument, which is used in order to bound the second largest eigenvalue of the adjacency matrices of random regular graphs \cite{FKS1989, FO2005, CGJ2018}. Similar approach is emerged in \cite{ALS2018} for bounding the spectral norm of $\mathbf{A}^{0}$, a \textit{processed similarity matrix}, which is obtained by zeroing-out every row and column of $\mathbf{A}$ whose sum is larger than a certain threshold. More precisely, $\mathbf{A}^{0}$ is obtained by zeroing-out both the $i^{\textnormal{th}}$ row and column of $\mathbf{A}$ if $\sum_{j=1}^{n} \mathbf{A}_{ij} > c_{\textnormal{thr}} \cdot \frac{1}{n} \sum_{i=1}^{n} \sum_{j=1}^{n} \mathbf{A}_{ij}$, where $c_{\textnormal{thr}} > 0$ is a threshold constant. We do not proceed such a trimming step and can still prove a concentration bound directly for $\mathbf{A}$, tighter than one obtained from matrix Bernstein inequality by a logarithmic factor. Moreover, our result does not assume any block structure in the underlying model, but only assumes that a random weight $W_e \in [0, 1]$ is independently assigned to each $e \in \mathcal{E}$ and the maximum expectation of weights is bounded as $\max \left\{ \mathbb{E}[W_e] : e \in \mathcal{E} \right\} \leq \mu_n$ with $n \binom{n-2}{d-2} \mu_n = \Omega ( \log n )$.

\begin{thm} 
\label{thm7.1}
Suppose that a random weight $W_e \in [0, 1]$ is independently assigned to each $d$-regular edge $e \in \mathcal{E}$,  where $\mathcal{H} = \left( [n], \left( W_e, e \in \mathcal{E} \right) \right)$ is a weighted $d$-uniform random hypergraph, and let $\mathbf{A}$ be the similarity matrix of $\mathcal{H}$. Also, we assume that $\max \left\{ \mathbb{E}[W_e] : e \in \mathcal{E} \right\} \leq \mu_n$, where $\left\{ \mu_n \right\}$ is a sequence in $(0, \infty)$ such that $n \binom{n-2}{d-2} \mu_n \geq \alpha \log n $ for some constant $\alpha > 0$. Then, there is a universal constant $c_5 > 0$ (depending on $\alpha$) such that with probability at least $1 - 4n^{-11}$, the similarity matrix $\mathbf{A}$ obeys the spectral norm bound
\beq
    \label{eq16}
    \left\| \mathbf{A} - \mathbb{E}[\mathbf{A}] \right\| \leq c_5 \sqrt{n \binom{n-2}{d-2} \mu_n}.
\eeq
\end{thm}

\indent The proof of Theorem \ref{thm7.1} is deferred to Appendix \ref{appendix:b}. We remark that the extra $\sqrt{\log n}$ factor from the bound obtained by matrix Bernstein inequality was removed in \eqref{eq16}. Also note that our bound \eqref{eq16} is a generalization of \emph{Theorem 5.2} in \cite{LR2015}, which provides a sharp spectral bound of the adjacency matrix for the \textit{graph case}. By using Theorem \ref{thm7.1}, we can obtain the corresponding result for the assortative $d\textsf{-WHSBM}$ directly.

\begin{cor} 
\label{cor7.1}
Let $\mathbf{A}$ denote the similarity matrix of $\mathcal{H} = \left( [n], \mathbf{W} \right)$ sampled by $d\textnormal{\textsf{-WHSBM}}(n, k, p_n, q_n, \Phi^*)$ with $p_n > q_n$. Suppose that there is an absolute constant $\alpha > 0$ such that $n \binom{n-2}{d-2} p_n \geq \alpha \log n$. Then, there is an absolute constant $c_5 > 0$ (depending on $\alpha$) such that with probability exceeding $1 - 4n^{-11}$, the similarity matrix $\mathbf{A}$ satisfies the bound
\beq
    \label{eq17}
        \left\| \mathbf{A} - \mathbb{E} [\mathbf{A}] \right\| \leq c_5 \sqrt{n \binom{n-2}{d-2} p_n}.
\eeq
\end{cor}

\subsection{Proof of Theorem \ref{thm4.1}} 
\label{proof_thm4.1}

\indent We first introduce additional notations and settings that will be needed in the proof of Theorem \ref{thm4.1}. Let $\nu_n := \binom{n-2}{d-2} p_n \geq \max \left\{ \mathbb{E}[\mathbf{A}_{ij}] : i, j \in [n] \right\}$. Define the \textit{normalized membership matrix} corresponding to the ground-truth community assignment $\Phi^*$ by a matrix $\mathbf{U} \in \mathbb{R}^{n \times k}$:
\begin{equation*}
    \mathbf{U}_{ia} :=
    \begin{cases}
        \frac{1}{\sqrt{s_a}} & \textnormal{if } \Phi^*(i) = a; \\
        0 & \textnormal{otherwise}.
    \end{cases}
\end{equation*}
\noindent Let $T$ be the linear subspace of $\mathbb{R}^{n \times n}$ spanned by elements of the form $\mathbf{U}_{*a} \cdot x^{\top}$ and $y \cdot \mathbf{U}_{*a}^{\top}$ for $a \in [k]$, where $x$ and $y$ are arbitrary vectors in $\mathbb{R}^n$, and $T^{\perp}$ be its orthogonal complement. The subspace $T$ of $\mathbb{R}^{n \times n}$ can be expressed explicitly by $ T = \left\{ \mathbf{U} \mathbf{A}^{\top} + \mathbf{B} \mathbf{U}^{\top} : \mathbf{A}, \mathbf{B} \in \mathbb{R}^{n \times k} \right\}$. The orthogonal projection $\mathcal{P}_{T}$ onto $T$ is given by $\mathcal{P}_T(\mathbf{X}) = \mathbf{U} \mathbf{U}^{\top} \mathbf{X} + \mathbf{X} \mathbf{U} \mathbf{U}^{\top} - \mathbf{U} \mathbf{U}^{\top} \mathbf{X} \mathbf{U} \mathbf{U}^{\top}$ and the orthogonal projection $\mathcal{P}_{T^{\perp}}$ onto $T^{\perp}$ is given by $\mathcal{P}_{T^{\perp}}(\mathbf{X}) = \left( \mathcal{I} - \mathcal{P}_{T} \right) (\mathbf{X}) = \left( \mathbf{I}_n - \mathbf{U} \mathbf{U}^{\top} \right) \mathbf{X} \left( \mathbf{I}_n - \mathbf{U} \mathbf{U}^{\top} \right)$. \\
\indent Recall that $\mathcal{X} \subseteq \mathbb{R}^{n \times n}$ refers to the feasible region of the semi-definite program \eqref{eq9}. To prove Theorem \ref{thm4.1}, it suffices to show that for any $\mathbf{X} \in \mathcal{X} \setminus \left\{ \mathbf{X}^* \right\}$,
\begin{equation*}
    \Delta(\mathbf{X}) := \left\langle \mathbf{A} - \lambda \mathbf{1}_{n \times n}, \mathbf{X}^* - \mathbf{X} \right\rangle > 0.
\end{equation*}
Using the orthogonal projections $\mathcal{P}_{T}$ and $ \mathcal{P}_{T^{\perp}}$, we propose to decompose the quantity $\Delta(\mathbf{X})$ as
\begin{equation*}
    \begin{split}
    \Delta(\mathbf{X}) &= \underbrace{\left\langle \mathcal{P}_{T} \left( \mathbf{A} - \mathbb{E}[\mathbf{A}] \right), \mathbf{X}^* - \mathbf{X} \right\rangle}_{\textnormal{(Q1)}} \\
    &+\underbrace{\left\langle \mathcal{P}_{T^{\perp}} \left( \mathbf{A} - \mathbb{E}[\mathbf{A}] \right), \mathbf{X}^* - \mathbf{X} \right\rangle}_{\textnormal{(Q2)}} \\
    &+\underbrace{\left\langle \mathbb{E}[\mathbf{A}] - \lambda \mathbf{1}_{n \times n}, \mathbf{X}^* - \mathbf{X} \right\rangle}_{\textnormal{(Q3)}}.
    \end{split}
\end{equation*}
The subsequent arguments on bounding the terms (Q1), (Q2) and (Q3) is akin to ones in \cite{CX2016, CLX2018}, except that the entries of $\mathbf{A}$ are not independent anymore for the hypergraph case so that it requires our new spectral bound (Corollary \ref{cor7.1}). Choose a sufficiently large constant $c_1 > 1$ and a universal constant $c_5 = c_5(1) > 0$ which is specified in Corollary \ref{cor7.1} with $\alpha = 1$. Below we establish lower bounds on the terms (Q1), (Q2) and (Q3):

\subsubsection{Lower bound of $\textnormal{(Q1)}$} 
The following lemma provides a sharp concentration bound on the $l_{\infty}$ norm of $\mathcal{P}_{T} \left( \mathbf{A} - \mathbb{E}[\mathbf{A}] \right)$:

\begin{lem}
\label{lem7.1}
Under the $d\textnormal{\textsf{-WHSBM}}(n, k, p_n, q_n, \Phi^*)$ with $p_n > q_n$ satisfying the condition \eqref{eq14}, the following bound holds with probability at least $1-2n^{-11}$\textnormal{:}
\beq
    \label{eq18}
    \begin{split}
    &\left\| \mathcal{P}_{T} \left( \mathbf{A} - \mathbb{E}[\mathbf{A}] \right) \right\|_{\infty}\\
    &\leq 3(d-1) \left( \frac{9}{c_1} + \sqrt{\frac{26}{c_1}} \right) \binom{s_{\min}-2}{d-2}(p_n - q_n).
    \end{split}
\eeq
\end{lem}

\indent The detailed proof of Lemma \ref{lem7.1} is deferred to Appendix \ref{appendix:c.1}. Invoking Lemma \ref{lem7.1} and H\"{o}lder's inequality together give with probability exceeding $1-2n^{-11}$,
\beq
    \label{eq19}
    \begin{split}
    \textnormal{(Q1)} \geq& - \left\| \mathcal{P}_{T} \left( \mathbf{A} - \mathbb{E}[\mathbf{A}] \right) \right\|_{\infty} \cdot \left\| \mathbf{X}^* - \mathbf{X} \right\|_{1} \\
    \geq& -3(d-1) \left( \frac{9}{c_1} + \sqrt{\frac{26}{c_1}} \right) \binom{s_{\min}-2}{d-2} \\
    &(p_n - q_n)\left\| \mathbf{X}^* - \mathbf{X} \right\|_{1}.
    \end{split}
\eeq

\subsubsection{Lower bound of $\textnormal{(Q2)}$} 
The ground-truth cluster matrix $\mathbf{X}^* = \mathbf{Z}^* \left( \mathbf{Z}^* \right)^{\top}$ has a rank-$k$ SVD given by $\mathbf{X}^* = \mathbf{U} \Sigma \mathbf{U}^{\top}$, where $\Sigma \in \mathbb{R}^{k \times k}$ is the diagonal matrix with entries $\Sigma_{aa} = s_a$, $a \in [k]$. Then, the sub-differential of the nuclear norm $\| \cdot \|_{*}$ at $\mathbf{X}^*$ is expressed as
\beq
    \label{eq20}
    \partial \left\| \mathbf{X}^* \right\|_{*} = \left\{ \mathbf{B} \in \mathbb{R}^{n \times n} : \mathcal{P}_{T}(\mathbf{B}) = \mathbf{U} \mathbf{U}^{\top}, \left\| \mathcal{P}_{T^{\perp}}(\mathbf{B}) \right\| \leq 1 \right\}.
\eeq
\noindent See \textit{Example 2} in \cite{W1992} or \cite{RFP2010} for characterization of sub-gradients of the nuclear norm. Therefore, $\mathcal{P}_{T^{\perp}} \left( \frac{\mathbf{A} - \mathbb{E}[\mathbf{A}]}{\left\| \mathbf{A} - \mathbb{E}[\mathbf{A}] \right\|} \right) + \mathbf{U} \mathbf{U}^{\top} \in \partial \left\| \mathbf{X}^* \right\|_{*}$ from \eqref{eq20}. It follows that for any $\mathbf{X} \in \mathcal{X}$,
\beq
    \label{eq21}
    \begin{split}
        0 &= \textnormal{Trace}(\mathbf{X}) - \textnormal{Trace}(\mathbf{X}^*) 
        \overset{\textnormal{(a)}}{=} \left\| \mathbf{X} \right\|_{*} - \left\| \mathbf{X}^* \right\|_{*} \\
        &\overset{\textnormal{(b)}}{\geq} \left\langle \mathbf{U} \mathbf{U}^{\top} + \mathcal{P}_{T^{\perp}} \left( \frac{\mathbf{A} - \mathbb{E}[\mathbf{A}]}{\left\| \mathbf{A} - \mathbb{E}[\mathbf{A}] \right\|} \right), \mathbf{X} - \mathbf{X}^* \right\rangle,
    \end{split}
\eeq
where the step (a) holds since both $\mathbf{X}$ and $\mathbf{X}^*$ are positive semi-definite matrices and (b) follows from the definition of sub-gradient. Hence, we obtain a lower bound on (Q2),
\beq
    \label{eq22}
    \begin{split}
        \textnormal{(Q2)} &= \left\langle \mathcal{P}_{T^{\perp}}(\mathbf{A} - \mathbb{E}[\mathbf{A}]), \mathbf{X}^* - \mathbf{X} \right\rangle\\
        &\overset{\textnormal{(c)}}{\geq} - \left\| \mathbf{A} - \mathbb{E}[\mathbf{A}] \right\| \cdot \left\langle \mathbf{U} \mathbf{U}^{\top}, \mathbf{X}^* - \mathbf{X} \right\rangle \\
        &\overset{\textnormal{(d)}}{\geq} - \left\| \mathbf{A} - \mathbb{E}[\mathbf{A}] \right\| \cdot \left\| \mathbf{U} \mathbf{U}^{\top} \right\|_{\infty} \cdot \left\| \mathbf{X}^* - \mathbf{X} \right\|_{1}\\
        &\overset{\textnormal{(e)}}{\geq} - \frac{1}{s_{\min}} \left\| \mathbf{A} - \mathbb{E}[\mathbf{A}] \right\| \cdot \left\| \mathbf{X}^* - \mathbf{X} \right\|_{1},
    \end{split}
\eeq 
where (c) is a consequence of \eqref{eq21}, (d) holds by the H\"{o}lder's inequality, and the step (e) is due to the fact that $\left[ \mathbf{U} \mathbf{U}^{\top} \right]_{ij} = 1 / s_{\Phi^{*}(i)} \leq 1 / s_{\min}$ whenever $\Phi^*(i) = \Phi^*(j)$; 0 otherwise. Also, the condition \eqref{eq14} leads to the inequality $n \binom{n-2}{d-2} p_n \geq \log n$, $\forall n \in \mathbb{N}$. Thus, the centered similarity matrix obeys the following bound with probability at least $1 - 4n^{-11}$:
\beq
    \label{eq23}
        \begin{split}
        \left\| \mathbf{A} - \mathbb{E}[\mathbf{A}] \right\| &\overset{\textnormal{(f)}}{\leq}
        c_5 \sqrt{n \binom{n-2}{d-2} p_n} \\
        &\overset{\textnormal{(g)}}{\leq} c_5 \sqrt{\frac{1}{c_1}} \cdot s_{\min} \binom{s_{\min}-2}{d-2} \left( p_n - q_n \right),
    \end{split}
\eeq
where the step (f) follows from Corollary \ref{cor7.1} and (g) is due to the condition \eqref{eq14}. We thus conclude by using \eqref{eq22} and \eqref{eq23} that with probability exceeding $1 - 4n^{-11}$,
\begin{equation}
    \label{eq24}
    \textnormal{(Q2)} \geq - c_5 \sqrt{\frac{1}{c_1}} \binom{s_{\min}-2}{d-2} \left( p_n - q_n \right)\left\| \mathbf{X}^* - \mathbf{X} \right\|_{1}.
\end{equation}
\noindent Here, we choose a sufficiently large constant $c_1$ such that $3(d-1) \left( \frac{9}{c_1} + \sqrt{\frac{26}{c_1}} \right) + \frac{c_5}{\sqrt{c_1}} \leq \frac{1}{8}$.

\subsubsection{Lower bound of $\textnormal{(Q3)}$}
First notice that $\mathbb{E}[\mathbf{A}_{ij}] \geq p_{n}^{-}$ if $i \neq j$ and $\mathbf{X}_{ij}^* = 1$; $\mathbb{E}[\mathbf{A}_{ij}] \leq q_{n}^{+}$ if $i \neq j$ and $\mathbf{X}_{ij}^* = 0$ (Actually, $\mathbb{E}[\mathbf{A}_{ij}] = q_{n}^{+}$ for $i \neq j$ in $[n]$ with $\mathbf{X}_{ij}^* = 0$). Then,
\beq
    \label{eq25}
    \begin{split}
        \textnormal{(Q3)}
        &= \sum_{i, j \in [n] : i \neq j} \left( \mathbb{E}[\mathbf{A}_{ij}] - \lambda \right) \left( \mathbf{X}_{ij}^* - \mathbf{X}_{ij} \right) \\
        &\quad+ \sum_{i=1}^{n} \left( -\lambda \right) \underbrace{\left( \mathbf{X}_{ii}^* - \mathbf{X}_{ii} \right)}_{= \ 0} \\
        &\overset{\textnormal{(h)}}{\geq} \sum_{\substack{i \neq j: \\ \mathbf{X}_{ij}^* = 1}} \left( p_{n}^{-} - \lambda \right) \left( 1 - \mathbf{X}_{ij} \right) + \sum_{\substack{i \neq j: \\ \mathbf{X}_{ij}^* = 0}} \left( \lambda - q_{n}^{+} \right) \mathbf{X}_{ij} \\
        &\overset{\textnormal{(i)}}{\geq} \frac{1}{4} \left( p_{n}^- - q_{n}^+ \right) \sum_{i, j \in [n] : i \neq j} \left| \mathbf{X}_{ij}^* - \mathbf{X}_{ij} \right|\\
        &\overset{\textnormal{(j)}}{=} \frac{1}{4} \binom{s_{\min}-2}{d-2} \left( p_n - q_n \right) \left\| \mathbf{X}^* - \mathbf{X} \right\|_{1},
    \end{split}
\eeq
where both the step (h) and (j) are due to the fact $\mathbf{X}_{ii} = 1$, $\forall i \in [n]$ for any feasible $\mathbf{X}$, which is deduced from the constraints $\textnormal{Trace}(\mathbf{X}) = n$ and $\mathbf{X}_{ij} \in [0, 1]$, $\forall i, j \in [n]$, and (i) follows from the condition \eqref{eq13}. \\
\indent To sum up, combining the above bounds on (Q1), (Q2) and (Q3), we may conclude by a union bound that with probability at least $1 - 6n^{-11}$,
\begin{equation*}
    \begin{split}
        \Delta(\mathbf{X}) &= \textnormal{(Q1)} + \textnormal{(Q2)} + \textnormal{(Q3)}\\
        & \geq \frac{1}{8} \binom{s_{\min}-2}{d-2} \left( p_n - q_n \right) \left\| \mathbf{X}^* - \mathbf{X} \right\|_{1},
    \end{split}
\end{equation*}
thereby showing that $\Delta(\mathbf{X}) > 0$ for all $\mathbf{X} \in \mathcal{X} \setminus \left\{ \mathbf{X}^* \right\}$.

\section{Conclusion}
\label{conclusion}

\indent In this paper, we developed an efficient hypergraph clustering method (\textsf{CRTMLE}) on a basis of the truncate-and-relax strategy, and proved its strong consistency guarantee under the assortative $d$\textsf{-WHSBM} with growing number of communities of order-wise unbalanced sizes. Our results are consistent with state-of-the-art results in various parameter regimes of the model, and settle the first strong consistency result for the case in which there are multiple communities of unbalanced sizes with different orders. Also, we manifested the robustness of \textsf{CRTMLE} against the unbalancedness of cluster sizes and the presence of outlier nodes, both theoretically and empirically.

\appendices

\section{Proof of Theorem \ref{thm4.2}}
\label{appendix:a}

Let $\mathcal{F}$ denote the event that the spectral norm bound \eqref{eq17} in Corollary \ref{cor7.1} holds with $\alpha = 1$. Note that Corollary \ref{cor7.1} states that $\mathbb{P} \{ \mathcal{F} \} \geq 1 - 4n^{-11}$, since the condition \eqref{eq14} directly yields the inequality $n \binom{n-2}{d-2} p_n \geq \log n$. It then follows that, on the event $\mathcal{F}$,
\beq
    \label{eqa.1}
    \max \left\{ \left| \hat{\lambda}_{i} - \lambda_{i} \right| : i \in [n] \right\} 
    \overset{\textnormal{(a)}}{\leq} \left\| \mathbf{A} - \mathbb{E}[\mathbf{A}] \right\| \leq c_5 \sqrt{n \nu_n},
\eeq
where (a) holds by the Weyl's inequality \cite{B1987}, $\nu_n := \binom{n-2}{d-2} p_n$, and $c_5 = c_5(1) > 0$. We henceforth assume that we are on the event $\mathcal{F}$.

\subsection{Estimation of $k$ and $s$}
The triangle inequality and the bound \eqref{eqa.1} imply that
\beq
    \label{eqa.2}
    \begin{split}
        \hat{\lambda}_{i} - \hat{\lambda}_{i+1} &= \left( \hat{\lambda}_{i} - \lambda_i \right) - \left( \hat{\lambda}_{i+1} - \lambda_{i+1} \right)\\
        &\leq \left| \hat{\lambda}_{i} - \lambda_i \right| + \left| \hat{\lambda}_{i+1} - \lambda_{i+1} \right| \leq 2 c_5 \sqrt{n \nu_n}
    \end{split}
\eeq
for every $i \in [n-1] \setminus \left\{ 1, k \right\}$, and
\beq
    \label{eqa.3}
    \begin{split}
        \hat{\lambda}_{k} - \hat{\lambda}_{k+1} &\geq \lambda_{k} - \lambda_{k+1} - \left| \hat{\lambda}_{k} - \lambda_{k} \right| - \left| \hat{\lambda}_{k+1} - \lambda_{k+1} \right| \\
        &\overset{\textnormal{(b)}}{\geq} s \binom{s-2}{d-2} \left( p_n - q_n \right) - 2c_5 \sqrt{n \nu_n},
    \end{split}
\eeq
where the step (b) follows by \eqref{eq15}. Then, one has
\begin{equation*}
    \begin{split}
    n \nu_n = n \binom{n-2}{d-2} p_n
    &\overset{\textnormal{(c)}}{\leq} \frac{1}{c_2} s^2 \binom{s-2}{d-2}^2 \left( p_n - q_n \right)^2,
    \end{split}
\end{equation*}
where the step (c) is due to the condition \eqref{eq14}. Therefore, it's straightforward that
\beq
    \label{eqa.4}
    s \binom{s-2}{d-2} \left( p_n - q_n \right) \geq \sqrt{c_2} \sqrt{n \nu_n} > 8 c_5 \sqrt{n \nu_n},
\eeq
since $c_2 > 0$ is chosen sufficiently large. Putting \eqref{eqa.4} together with inequalities \eqref{eqa.2} and \eqref{eqa.3}, we have
\beq
    \label{eqa.5}
    \begin{split}
    \hat{\lambda}_{k} - \hat{\lambda}_{k+1}& \geq s \binom{s-2}{d-2} \left( p_n - q_n \right) - 2c_5 \sqrt{n \nu_n} \\
    &> 2c_5 \sqrt{n \nu_n} \geq \hat{\lambda}_{i} - \hat{\lambda}_{i+1}
\end{split}
\eeq
for all $i > 1$ with $i \neq k$. This guarantees $\hat{k} = k$ and $\hat{s} = s$.

\subsection{Estimation of $p_{n}^{-}$ and $q_{n}^{+}$}
From the triangle inequality and \eqref{eqa.1}, the estimation error of $\hat{p}_{n}^{-}$ obeys
\beq
    \label{eqa.6}
    \begin{split}
    \left| \hat{p}_{n}^{-} - p_{n}^{-} \right| &\overset{\textnormal{(d)}}{=} \left| \frac{s \left( \hat{\lambda}_{1} - \lambda_{1} \right) + (n-s) \left( \hat{\lambda}_{2} - \lambda_{2} \right)}{n (s-1)} \right| \\
    &\leq \frac{s}{n(s-1)} \left| \hat{\lambda}_{1} - \lambda_{1} \right| + \frac{n-s}{n(s-1)} \left| \hat{\lambda}_{2} - \lambda_{2} \right| \\
    &\leq 2c_5 \frac{\sqrt{n \nu_n}}{s},
    \end{split}
\eeq
where the step (d) is owing to $\hat{s} = s$. Similarly, we can deduce the following estimation error of $\hat{q}_{n}^{+}$:
\beq
    \label{eqa.7}
        \begin{split}
        \left| \hat{q}_{n}^{+} - q_{n}^{+} \right| &= \frac{1}{n} \left| \left( \hat{\lambda}_{1} - \lambda_{1} \right) - \left( \hat{\lambda}_{2} - \lambda_{2} \right) \right| \\
        &\leq
        2c_5 \sqrt{\frac{\nu_n}{n}} \leq 2c_5 \frac{\sqrt{n \nu_n}}{s}.
    \end{split}
\eeq

\subsection{Estimation of $\lambda$}
The error bounds \eqref{eqa.6} and \eqref{eqa.7} yield
\beq
    \label{eqa.8}
    \begin{split}
    \hat{\lambda} &= \frac{p_{n}^{-} + q_{n}^{+}}{2} + \frac{\hat{p}_{n}^{-} - p_{n}^{-}}{2} + \frac{\hat{q}_{n}^{+} - q_{n}^{+}}{2}\\
    &\leq \frac{p_{n}^{-} + q_{n}^{+}}{2} + 2c_5 \frac{\sqrt{n \nu_n}}{s} \\
    &\overset{\textnormal{(e)}}{\leq} \frac{p_{n}^{-} + q_{n}^{+}}{2} + \frac{1}{4} \binom{s-2}{d-2} \left( p_n - q_n \right)\\
    &= \frac{3}{4} p_{n}^{-} + \frac{1}{4} q_{n}^{+},
    \end{split}
\eeq
where the inequality (e) holds by \eqref{eqa.4}. The above bound \eqref{eqa.8} gives the desired upper bound on the estimator $\hat{\lambda}$ of $\lambda$. In a similar manner, we can derive the desired lower bound of $\hat{\lambda}$.

\section{Proof of Theorem \ref{thm7.1}}
\label{appendix:b}

\indent The proof of Theorem \ref{thm7.1} is technically involved as it is built upon a celebrated combinatorial argument, which is often utilized in order to derive spectral bounds for random matrices \cite{FKS1989, FO2005, CGJ2018, ALS2018, LR2015, CRV2015}. We begin the proof from a basic idea that we use to bound
\beq
    \label{eqb.0}
    \left\| \mathbf{A} - \mathbb{E}[\mathbf{A}] \right\| = \sup_{x \in \mathbb{S}^{n-1}} \left| \left\langle \left( \mathbf{A} - \mathbb{E}[\mathbf{A}] \right) x, x \right\rangle \right|.
\eeq
Here, we provide a brief overview of three major steps.
\begin{enumerate}
    \item \emph{Discretization of the unit $n$-sphere $\mathbb{S}^{n-1}$}: First, we reduce \eqref{eqb.0} to the problem of bounding the supremum of the quadratic form $\left\langle \left( \mathbf{A} - \mathbb{E}[\mathbf{A}] \right) x, x \right\rangle$ over $x \in \mathcal{N}$, where $\mathcal{N}$ is a finite subset of the unit $n$-sphere so that $\mathbb{S}^{n-1}$ can be covered by closed balls of the same radii with centers in $\mathcal{N}$. See Lemma \ref{lemb.1} for validation of such a manipulation. The quantity $\left\langle \left( \mathbf{A} - \mathbb{E}[\mathbf{A}] \right) x, x \right\rangle$ can be decomposed as the sum of two parts. The first part corresponds to the pairs $(i, j) \in [n] \times [n]$ such that $|x_i x_j|$ is small, called \emph{light couples}, while another part corresponds to the pairs such that $|x_i x_j|$ is large, the \emph{heavy couples}.
    \item \emph{Managing the light couples}: To control the contribution of light couples at each point of $\mathcal{N}$, we use the standard Bernstein's inequality. Then, we completes bounding the supremum over $\mathcal{N}$ by employing the union bound.
    \item \emph{Managing the heavy couples}: Finally, in order to manipulate the contribution of heavy couples, we show that the similarity matrix $\mathbf{A}$ has the \emph{discrepancy property}, which essentially says that the sum of edge weights between any two subsets of vertices does not deviate much from its expectations. Then, we use the fact that the discrepancy property of a given matrix ensures a small contribution of the heavy couples to the quadratic form.
\end{enumerate}

\begin{lem} 
\label{lemb.1}
Suppose that $0 < \varepsilon < \frac{1}{2}$ and $\mathbf{M}$ is an $n \times n$ real symmetric matrix. Then, for any $\varepsilon$-net $\mathcal{N}$ on the unit $n$-sphere $\mathbb{S}^{n-1}$, we have the bound
\beq
    \label{eqb.1}
    \begin{split}
    &\sup \left\{ \left| \langle \mathbf{M}x, x \rangle \right| : x \in \mathcal{N} \right\} \\
    \leq\ &\| \mathbf{M} \|
    \leq \frac{1}{1 - 2 \varepsilon} \sup \left\{ \left| \langle \mathbf{M}x, x \rangle \right| : x \in \mathcal{N} \right\}.
    \end{split}
\eeq
\end{lem}

\indent The detailed proof of Lemma \ref{lemb.1} is deferred to Appendix \ref{appendix:c.2}. On the other hand, some volumetric arguments yield the following result.

\begin{lem} [\textit{Corollary 4.2.13} in \cite{V2018}; \textit{Lemma 6.8} in \cite{CGJ2018}]
\label{lemb.2}
Let $E \subset \mathbb{S}^{n-1}$ be a subset of the unit $n$-sphere and $\varepsilon > 0$. Then, there exists an $\varepsilon$-net $\mathcal{N}$ of $E$ such that $\left| \mathcal{N} \right| \leq \left( 1 + \frac{2}{\varepsilon} \right)^n$.
\end{lem}

\indent Now, we take $\varepsilon = \frac{1}{4}$ and apply Lemma \ref{lemb.2}. It guarantees the existence of an $\frac{1}{4}$-net $\mathcal{N}$ of $\mathbb{S}^{n-1}$ with $\left| \mathcal{N} \right| \leq 9^n$. Also, the following inequality holds due to Lemma \ref{lemb.1}:
\beq
    \label{eqb.2}
    \| \mathbf{A} - \mathbb{E}[\mathbf{A}] \| \leq 2 \sup \left\{ \left| x^{\top} ( \mathbf{A} - \mathbb{E}[\mathbf{A}] ) x \right| : x \in \mathcal{N} \right\}.
\eeq

\noindent Thus, it suffices to upper-bound the RHS of \eqref{eqb.2}. Before we elaborate the bounding argument of the RHS, we delineate a key step in the Friedman-Kahn-Szemer\'{e}di argument, which is the separation of $x^{\top} \mathbf{A} x = \sum_{i=1}^{n} \sum_{j=1}^{n} x_i x_j \mathbf{A}_{ij}$ into two pieces. First, we write $\theta_n := \binom{n-2}{d-2} \mu_n \geq \max_{i, j \in [n]} \mathbb{E}[\mathbf{A}_{ij}]$. Given any $x \in \mathbb{S}^{n-1}$, the \textit{light couples} and the \textit{heavy couples} of nodes are given by
\begin{equation*}
    \begin{split}
    \mathcal{L}(x) &:= \left\{ (i, j) \in [n] \times [n] : \left| x_i x_j \right| \leq \sqrt{\frac{\theta_n}{n}} \right\} \textnormal{ and} \\
    \mathcal{H}(x) &:= \left( [n] \times [n] \right) \setminus \mathcal{L}(x),
    \end{split}
\end{equation*}
respectively, and the $n \times n$ real matrices $L(x)$ and $H(x)$ are given by $\left[ L(x) \right]_{ij} = x_i x_j$ if $(i, j) \in \mathcal{L}(x)$; 0 otherwise, and $H(x) := xx^{\top} - L(x)$. Fix any point $x \in \mathcal{N}$ and apply the triangle inequality to obtain
\beq
    \label{eqb.3}
    \begin{split}
        &\sup_{x \in \mathcal{N}} \left|  x^{\top} \left( \mathbf{A} - \mathbb{E}[\mathbf{A}] \right) x \right| \leq \\
        &\quad\underbrace{\sup_{x \in \mathcal{N}} \left| \sum_{(i,j) \in \mathcal{L}(x)} x_i x_j \mathbf{A}_{ij} - x^{\top} \mathbb{E}[\mathbf{A}] x \right|}_{\textnormal{(T1)}} \\
        &\quad+ \underbrace{\sup_{x \in \mathcal{N}} \left| \sum_{(i,j) \in \mathcal{H}(x)} x_i x_j \mathbf{A}_{ij} \right|}_{\textnormal{(T2)}}.
    \end{split}
\eeq

\subsection{Bound of \textnormal{(T1)}} By the triangle inequality, we have
\begin{equation*}
    \begin{split}
        \textnormal{(T1)} &\leq \underbrace{\sup_{x \in \mathcal{N}} \left| \sum_{(i,j) \in \mathcal{L}(x)} x_i x_j \left( \mathbf{A}_{ij} - \mathbb{E}[\mathbf{A}_{ij}] \right) \right|}_{\textnormal{(E1)}} \\
        &+ \underbrace{\sup_{x \in \mathcal{N}} \left| \sum_{(i, j) \in \mathcal{H}(x)} x_i x_j \mathbb{E}[\mathbf{A}_{ij}] \right|}_{\textnormal{(E2)}}.
    \end{split}
\end{equation*}
\begin{enumerate}
    \item (E1): For every $x \in \mathcal{N}$, we have the identity
    \begin{equation*}
        \begin{split}
            &\sum_{(i,j) \in \mathcal{L}(x)} x_i x_j \left( \mathbf{A}_{ij} - \mathbb{E}[\mathbf{A}_{ij}] \right) \\
            =& \sum_{e \in \mathcal{E}} \underbrace{\left( W_e - \mathbb{E}[W_e] \right) \left( \sum_{\substack{(i,j) \in \mathcal{L}(x): \\ i \neq j,\ \{i,j\} \subseteq e}} x_i x_j \right)}_{Y_e},
        \end{split}
    \end{equation*}
    Then, $Y_e$, $e \in \mathcal{E}$, are independent and mean-zero random variables. To apply the standard Bernstein's inequality, we need the following observations.
    \begin{itemize}
        \item $\left| Y_e \right| \leq \sum_{\substack{(i,j) \in \mathcal{L}(x): \\ i \neq j,\ \{ i,j \} \subseteq e}} \left| x_i x_j \right| \leq d^2 \sqrt{\frac{\theta_n}{n}}$ from the definition of light couples.
        \item An upper-bound on the sum of second moments of $Y_e$'s can be computed as:
        \begin{equation*}
            \begin{split}
                \sum_{e \in \mathcal{E}} \mathbb{E}[Y_{e}^2] &= \sum_{e \in \mathcal{E}} \textnormal{Var}[W_e] \left( \sum_{\substack{(i, j) \in \mathcal{L}(x): \\ i \neq j,\ \{i,j\} \subseteq e}} x_i x_j \right)^2 \\
                &\overset{\textnormal{(a)}}{\leq} d^2  \sum_{e \in \mathcal{E}} \sum_{\substack{(i, j) \in [n] \times [n]: \\ i \neq j,\ \{i,j\} \subseteq e}} \mathbb{E}[W_e] x_{i}^2 x_{j}^2 \\
                &\overset{\textnormal{(b)}}{\leq} d^2 \theta_n \cdot \sum_{(i,j) \in [n] \times [n]: i \neq j} x_{i}^2 x_{j}^2 \\
                &\leq d^2 \theta_n \left( \sum_{i=1}^{n} x_{i}^2 \right)^2 = d^2 \theta_n,
            \end{split}
        \end{equation*}
        where the step (a) is due to Cauchy-Schwarz inequality and the step (b) follows from the property that $W_e \in [0,1]$ for all $e \in \mathcal{E}$.
    \end{itemize}
    By the two-sided Bernstein's inequality, we have
    \begin{equation*}
        \mathbb{P} \left\{ \left| \sum_{e \in \mathcal{E}} Y_e \right| > \beta_1 \sqrt{n \theta_n} \right\} \leq 2 \exp \left( - \frac{\frac{1}{2} \beta_{1}^2}{d^2 \left( 1 + \frac{\beta_1}{3} \right)} n \right)
    \end{equation*}
    for any constant $\beta_1 > 0$. The union bound yields
    \begin{equation*}
        \begin{split}
        &\ \mathbb{P} \left\{ \textnormal{(E1)} > \beta_1 \sqrt{n \theta_n} \right\} \\
        \leq& \ | \mathcal{N} | \cdot 2 \exp \left( - \frac{\beta_{1}^2}{2d^2 \left( 1 + \frac{\beta_1}{3} \right)} n \right) \\
        \leq& \ 2 \exp \left\{ \left( 2 \log 3 - \frac{\beta_{1}^2}{2d^2 \left( 1 + \frac{\beta_1}{3} \right)} \right) n \right\}.
        \end{split}
    \end{equation*}
    So, if we choose a constant $\beta_1 > 0$ such that $2 \log 3 - \frac{\beta_{1}^2}{2d^2 \left( 1 + \frac{\beta_1}{3} \right)} \leq - 11$, then with prob. at least $1 - 2e^{-11n}$, we have $\textnormal{(E1)} \leq \beta_1 \sqrt{n \theta_n}$.
        
    \item (E2): From the definition of heavy couples, the following inequalities hold for any $x \in \mathbb{S}^{n-1}$,
    \begin{equation*}
        \begin{split}
            &\left| \sum_{(i, j) \in \mathcal{H}(x)} x_i x_j \mathbb{E}[\mathbf{A}_{ij}] \right| \leq
            \sum_{(i, j) \in \mathcal{H}(x)} \mathbb{E}[\mathbf{A}_{ij}] \frac{x_{i}^2 x_{j}^2}{\left| x_i x_j \right|}\\
            \leq& \ \sqrt{n \theta_n}\sum_{(i, j) \in \mathcal{H}(x)} x_{i}^2 x_{j}^2\leq \sqrt{n \theta_n}.
        \end{split}
    \end{equation*}
    Hence, it's clear that 
    \begin{equation*}
        \textnormal{(E2)} = \sup_{x \in \mathcal{N}} \left| \sum_{(i, j) \in \mathcal{H}(x)} x_i x_j \mathbb{E}[\mathbf{A}_{ij}] \right| \leq \sqrt{n \theta_n}.
    \end{equation*}
\end{enumerate}
    
\noindent By combining above two bounds together, we can ensure that with probability exceeding $1 - 2n^{-11}$,
\beq
    \label{eqb.4}
    \textnormal{(T1)} \leq \textnormal{(E1)} + \textnormal{(E2)} \leq \left( \beta_1 + 1 \right) \sqrt{n \theta_n}.
\eeq

\subsection{Bound of \textnormal{(T2)}} 
\indent Before we elaborate the bounding argument of (T2), we organize some preliminaries.
\begin{itemize}
    \item Let $h : (-1, \infty) \rightarrow \mathbb{R}$ be a function defined by $h(x) := (1+x) \log (1+x) - x$.
    \item Given any two matrices $\mathbf{A}, \mathbf{B} \in \mathbb{R}^{m \times n}$, the {Hadamard product} $\mathbf{A} \circ \mathbf{B}$ of $\mathbf{A}$ and $\mathbf{B}$ is defined as an $m \times n$ matrix with entries $\left[ \mathbf{A} \circ \mathbf{B} \right]_{ij} = \mathbf{A}_{ij} \mathbf{B}_{ij}$ for $i \in [m]$ and $j \in [n]$.
    \item For any $m \times n$ matrix $\mathbf{M}$ and $S \subset [m]$, $T \subset [n]$, we define $e_{\mathbf{M}}(S, T) := \sum_{i \in S} \sum_{j \in T} \mathbf{M}_{ij}$.
    \item For any matrix $\mathbf{Q} \in \mathbb{R}^{n \times n}$, we define $f_{\mathbf{Q}} : \mathbb{R}^{n \times n} \rightarrow \mathbb{R}$ by $f_{\mathbf{Q}}(\mathbf{X}) := \langle \mathbf{Q}, \mathbf{X} \rangle = \sum_{i, j \in [n]} \mathbf{Q}_{ij} \mathbf{X}_{ij}$.
\end{itemize}
    
\noindent Now, we summarize some concentration properties of random symmetric matrices that are used in the celebrated Friedman-Kahn-Szemer\'{e}di argument.

\begin{defi} [Uniform Upper Tail Property \cite{CGJ2018}] \normalfont{
\label{defib.1}
Let $\mathbf{M}$ be an $n \times n$ random symmetric matrix with non-negative entries. With the linear map $f_{\mathbf{Q}} : \mathbb{R}^{n \times n} \rightarrow \mathbb{R}$ defined above, we write
\begin{equation*}
    \begin{split}
    \mu &:= \mathbb{E} \left[ f_{\mathbf{Q}}(\mathbf{M}) \right] = f_{\mathbf{Q}} \left( \mathbb{E}[\mathbf{M}] \right);\ \\
    \tilde{\sigma}^2 &:= \mathbb{E} \left[ f_{\mathbf{Q} \circ \mathbf{Q}}(\mathbf{M}) \right] = f_{\mathbf{Q} \circ \mathbf{Q}} \left( \mathbb{E}[\mathbf{M}] \right).
    \end{split}
\end{equation*}
The matrix $\mathbf{M}$ is said to have the \textit{uniform upper tail property} $\textnormal{UUTP}(c_0, \gamma_0)$ with parameters $c_0 > 0$ and $\gamma_0 \geq 0$ if the following holds: for any $a, t > 0$ and $n \times n$ symmetric matrix $\mathbf{Q}$ with $\mathbf{Q}_{ij} \in [0, a]$, $\forall i, j \in [n]$, we have
\begin{equation*}
    \mathbb{P} \left\{ f_{\mathbf{Q}}(\mathbf{M}) - \mu \geq \gamma_0 \mu + t \right\} \leq \exp \left\{ - c_0 \frac{\tilde{\sigma}^2}{a^2} h \left( \frac{at}{\tilde{\sigma}^2} \right) \right\}.
\end{equation*}
}
\end{defi}

\begin{defi} [Discrepancy Property \cite{FO2005,CGJ2018}] \normalfont{
\label{defia.2}
 Let $\mathbf{M}$ be an $n \times n$ real matrix with non-negative entries. We say that $\mathbf{M}$ obeys the \textit{discrepancy property} $\textnormal{DP}(\delta, \kappa_1, \kappa_2)$ with parameters $\delta > 0$, $\kappa_1 > 0$ and $\kappa_2 \geq 0$ if for all non-empty $S, T \subseteq [n]$, at least one of the following properties hold:
\begin{enumerate} [label=(\alph*)]
    \item $e_{\mathbf{M}}(S, T) \leq \kappa_1 \delta \cdot |S| |T|$;
    \item $\nonumber e_{\mathbf{M}}(S, T) \log \frac{e_{\mathbf{M}}(S, T)}{\delta \cdot |S| |T|} \leq \kappa_2 \left( |S| \vee |T| \right) \log \frac{en}{|S| \vee |T|}.$
\end{enumerate}
}
\end{defi}

\indent The following lemma says the uniform upper tail property of a real symmetric random matrix whose entries are non-negative implies the discrepancy property \emph{w.h.p.}.

\begin{lem} [\textit{Lemma 6.4} in \cite{CGJ2018}]
\label{lemb.3}
Let $\mathbf{M}$ be an $n \times n$ symmetric random matrix with non-negative entries. Suppose that $\mathbb{E}[\mathbf{M}_{ij}] \leq \delta$, $\forall i, j \in [n]$ for some $\delta > 0$ and $\mathbf{M}$ obeys the $\textnormal{UUTP}(c_0, \gamma_0)$ for some $c_0 > 0$, $\gamma_0 \geq 0$. Then, for any $K > 0$, $\mathbf{M}$ satisfies the $\textnormal{DP}(\delta, \kappa_1, \kappa_2)$ with probability at least $1 - n^{-K}$, where $\kappa_1 = \kappa_1(c_0, \gamma_0, K) := e^2 (1 + \gamma_0)^2$ and $\kappa_2 = \kappa_2(c_0, \gamma_0, K) := \frac{4}{c_0} (1 + \gamma_0)(K+4)$.
\end{lem}

\indent Due to the ensuing result, whenever the similarity matrix $\mathbf{A}$ has the discrepancy property, we can show that the contribution of the heavy couples to \eqref{eqb.3} is not significant.

\begin{lem} [\textit{Lemma 6.6} in \cite{CGJ2018}] 
\label{lemb.4}
Let $\mathbf{M}$ be an $n \times n$ symmetric matrix with non-negative entries such that $\sum_{j=1}^{n} \mathbf{M}_{ij} \leq \lambda$, $\forall i \in [n]$ and $\sum_{i=1}^{n} \mathbf{M}_{ij} \leq \lambda$, $\forall j \in [n]$. Suppose $\mathbf{M}$ obeys the $\textnormal{DP}(\delta, \kappa_1, \kappa_2)$ with $\delta = \frac{C \lambda}{n}$, for some constants $C, \kappa_1 > 0$ and $\kappa_2 \geq 0$. Then for any $x \in \mathbb{S}^{n-1}$,
\begin{equation*}
    \left| f_{H(x)}(\mathbf{M}) \right| = \left| \sum_{(i, j) \in \mathcal{H}(x)} x_i x_j \mathbf{M}_{ij} \right| \leq \sigma(C, \kappa_1, \kappa_2) \sqrt{\lambda},
\end{equation*}
where $\sigma(C, \kappa_1, \kappa_2) := 64 \kappa_2 \left( 1 + \frac{2}{\kappa_1 \log \kappa_1} \right) + 32C(1 + \kappa_1) + 16$.
\end{lem}

\indent We claim that $\textnormal{(T2)} = \sup_{x \in \mathcal{N}} \left| f_{H(x)} (\mathbf{A}) \right| = O(\sqrt{n \theta_n})$ with probability at least $1 - 2n^{-11}$. Its proof consists of two main stages, which we elaborate below.

\subsubsection{The similarity matrix $\mathbf{A}$ obeys the uniform upper tail property}
First, we fix any $a, t > 0$ and any $n \times n$ symmetric matrix $\mathbf{Q}$ with entries $\mathbf{Q}_{ij} \in [0, a]$ for all $i, j$. Recall that $\mu_{\mathbf{Q}} := \mathbb{E}[f_{\mathbf{Q}}(\mathbf{A})]$ and $\tilde{\sigma}_{\mathbf{Q}}^2 := \mathbb{E}[f_{\mathbf{Q} \circ \mathbf{Q}}(\mathbf{A})]$. Then, we get
\begin{equation*}
    \begin{split}
    &\ f_{\mathbf{Q}}(\mathbf{A}) - \mu_{\mathbf{Q}} = \sum_{i, j \in [n] : i \neq j} \mathbf{Q}_{ij} \left( \mathbf{A}_{ij} - \mathbb{E}[\mathbf{A}_{ij}] \right) \\
    =&\ \sum_{e \in \mathcal{E}} \underbrace{\left( W_e - \mathbb{E}[W_e] \right) \cdot \left( \sum_{i, j \in e : i \neq j} \mathbf{Q}_{ij} \right)}_{Z_e}.
    \end{split}
\end{equation*}
Note that $Z_e$, $e \in \mathcal{E}$, are independent and mean-zero. We then use the Bennett's inequality on the sum $\sum_{e \in \mathcal{E}} Z_e$ by involving the following preliminary calculations:
\begin{itemize}
    \item $\left| Z_e \right| \leq \sum_{i, j \in e : i \neq j} \mathbf{Q}_{ij} \leq d^2 a$ for all $e \in \mathcal{E}$.
    \item Let $\sigma_{\mathbf{Q}}^2 := \sum_{e \in \mathcal{E}} \mathbb{E}[Z_{e}^2]$. Then, we obtain the bound
    \beq
        \label{eqb.5}
        \begin{split}
        \sigma_{\mathbf{Q}}^2 &= \sum_{e \in \mathcal{E}} \left( \sum_{i, j \in e : i \neq j} \mathbf{Q}_{ij} \right)^2 \textnormal{Var}[W_e] \\
        &\overset{\textnormal{(c)}}{\leq} \sum_{e \in \mathcal{E}} \left( \sum_{i, j \in e : i \neq j} \mathbf{Q}_{ij}^2 \right) \left( \sum_{i, j \in e : i \neq j} 1 \right) \textnormal{Var}[W_e] \\
        &\overset{\textnormal{(d)}}{\leq} d^2 \sum_{e \in \mathcal{E}} \left( \sum_{i, j \in e : i \neq j} \mathbf{Q}_{ij}^2 \right) \mathbb{E}[W_e]\\
        & = d^2 \sum_{i, j \in [n] : i \neq j} \mathbf{Q}_{ij}^2 \left( \sum_{e \in \mathcal{E} : \{i, j\} \subseteq e} \mathbb{E}[W_e] \right) = d^2 \cdot \tilde{\sigma}_{\mathbf{Q}}^2,
        \end{split}
    \eeq
    where the step (c) is due to the Cauchy-Schwarz inequality and (d) comes from the property $W_e \in [0, 1]$, $\forall e \in \mathcal{E}$.
\end{itemize}

\noindent The Bennett's inequality implies that for any $\gamma_0 \geq 0$, we have
\beq
    \begin{split}
    \label{eqb.6}
    & \ \mathbb{P} \left\{ f_{\mathbf{Q}}(\mathbf{A}) - \mu_{\mathbf{Q}} \geq \gamma_0 \mu_{\mathbf{Q}} + t \right\}\\
    \leq& \ \exp \left[ - \frac{\sigma_{\mathbf{Q}}^2}{\left( d^2 a \right)^2} h \left( \frac{d^2 a}{\sigma_{\mathbf{Q}}^2} \left( \gamma_0 \mu_{\mathbf{Q}} + t \right) \right) \right].
    \end{split}
\end{equation}
\noindent Meanwhile, the following bound holds for any $\gamma_0 \geq 0$,
\beq
    \label{eqb.7}
    \begin{split}
    \frac{\sigma_{\mathbf{Q}}^2}{\left( d^2 a \right)^2} h \left( \frac{d^2 a}{\sigma_{\mathbf{Q}}^2} \left( \gamma_0 \mu_{\mathbf{Q}} + t \right) \right) 
    &\overset{\textnormal{(e)}}{\geq} \frac{\sigma_{\mathbf{Q}}^2}{\left( d^2 a \right)^2} h \left( \frac{d^2 a}{\sigma_{\mathbf{Q}}^2} t \right) \\ &\overset{\textnormal{(f)}}{\geq} \frac{\tilde{\sigma}_{\mathbf{Q}}^2}{d^2 a^2} h \left( \frac{at}{\tilde{\sigma}_{\mathbf{Q}}^2} \right),
\end{split}
\eeq
where the step (e) holds by the observation that the function $h$ is non-decreasing on $[0, \infty)$ and (f) attributes to the fact that for fixed $a, t > 0$, the function $\lambda \in \left( 0, \infty \right) \mapsto \frac{\lambda}{d^4 a^2} h \left( \frac{d^2 a}{\lambda} t \right)$ is non-increasing and \eqref{eqb.5}. Combining the inequalities \eqref{eqb.6} and \eqref{eqb.7} together gives the bound
\begin{equation*}
    \begin{split}
    \mathbb{P} \left\{ f_{\mathbf{Q}}(\mathbf{A}) - \mu_{\mathbf{Q}} \geq \gamma_0 \mu_{\mathbf{Q}} + t \right\} \leq \exp \left[ - \frac{1}{d^2} \cdot \frac{\tilde{\sigma}_{\mathbf{Q}}^2}{a^2} h \left( \frac{at}{\tilde{\sigma}_{\mathbf{Q}}^2} \right) \right],
    \end{split}
\end{equation*}
which implies the $\textnormal{UUTP} \left( \frac{1}{d^2}, \gamma_0 \right)$ of the similarity matrix $\mathbf{A}$ for every $\gamma_0 \geq 0$. From now on, we set $\gamma_0 = 1$.
\subsubsection{All row and column sums of $\mathbf{A}$ is of order $O(n \theta_n)$ w.h.p.} Now, we will show that $\max \left\{ \sum_{i=1}^{n} \mathbf{A}_{ij} : j \in [n] \right\} = O(n \theta_n)$ \textit{w.h.p.}. We fix any $j \in [n]$ and observe that
\begin{equation*}
    \begin{split}
    \sum_{i=1}^{n} \mathbf{A}_{ij}
    &= \sum_{i \in [n] \setminus \{ j \}} \left( \sum_{e \in \mathcal{E} : \{ i, j \} \subseteq e} W_e \right)\\
    &= \sum_{e \in \mathcal{E}_j} \left( \sum_{i \in e \setminus \{ j \}} W_e \right) = (d-1) \sum_{e \in \mathcal{E}_j} W_e,
    \end{split}
\end{equation*}
where $\mathcal{E}_j := \left\{ e \in \mathcal{E} : j \in e \right\}$.  Translating both sides gives
\begin{equation*}
    \sum_{i=1}^{n} \left( \mathbf{A}_{ij} - \mathbb{E}[\mathbf{A}_{ij}] \right) = (d-1) \sum_{e \in \mathcal{E}_j} \left( W_e - \mathbb{E}[W_e] \right).
\end{equation*}
\noindent Here, we remark that
\begin{itemize}
    \item $\left| W_e - \mathbb{E}[W_e] \right| \leq 1$ for all $e \in \mathcal{E}_j$.
    \item We have the bound
    \begin{equation*}
        \begin{split}
        \sum_{e \in \mathcal{E}_j} \mathbb{E}[\left( W_e - \mathbb{E}[W_e] \right)^2] 
        &\overset{\textnormal{(g)}}{\leq} \sum_{e \in \mathcal{E}_j} \mathbb{E}[W_e]\\
        &\leq \mu_n \cdot \left| \mathcal{E}_j \right|
        = \frac{n \theta_n}{d-1},
        \end{split}
    \end{equation*}
    where (g) makes use of the property $W_e \in [0, 1]$, $\forall e \in \mathcal{E}_j$. 
\end{itemize}
Then, the one-sided Bernstein's inequality yields
\beq
    \label{eqb.8}
    \begin{split}
    &\ \mathbb{P} \left\{ \sum_{i=1}^{n} \left( \mathbf{A}_{ij} - \mathbb{E}[\mathbf{A}_{ij}] \right) > (d-1) \beta_{2} \cdot n \theta_n \right\}\\
    \leq& \ \exp \left( - \frac{\frac{1}{2} \beta_{2}^2}{\frac{\beta_2}{3} + \frac{1}{d-1}} n \theta_n \right) 
    \end{split}
\eeq
for any constant $\beta_2 > 0$. Since $n \theta_n \geq \alpha \log n$ for every $n \in \mathbb{N}$, the inequality \eqref{eqb.8} reduces to
\begin{equation*}
    \begin{split}
        &\ \mathbb{P} \left\{ \sum_{i=1}^{n} \left( \mathbf{A}_{ij} - \mathbb{E}[\mathbf{A}_{ij}] \right) > (d-1) \beta_{2} \cdot n \theta_n \right\}\\
        \leq& \ \exp \left( - \frac{\frac{1}{2} \alpha \beta_{2}^2}{\frac{\beta_2}{3} + \frac{1}{d-1}} \log n \right).
    \end{split}
\end{equation*}
By taking $\beta_2 > 0$ with $\frac{1}{2} \alpha \beta_{2}^2 \geq 12 \left( \frac{\beta_2}{3} + \frac{1}{d-1} \right)$, one has
\begin{equation*}
    \sum_{i=1}^{n} \left( \mathbf{A}_{ij} - \mathbb{E}[\mathbf{A}_{ij}] \right) \leq (d-1) \beta_2 \cdot n \theta_n
\end{equation*}
with probability exceeding $1 - n^{-12}$. Consequently, we have
\begin{equation*}
    \begin{split}
        \sum_{i=1}^{n} \mathbf{A}_{ij} &= \sum_{i=1}^{n} \left( \mathbf{A}_{ij} - \mathbb{E}[\mathbf{A}_{ij}] \right) + \sum_{i=1}^{n} \mathbb{E}[\mathbf{A}_{ij}] \\
        &\leq \left[ (d-1) \beta_2 + 1 \right] n \theta_n
    \end{split}
\end{equation*}
with probability greater than $1 - n^{-12}$, $\forall j \in [n]$. It follows by the union bound that with probability larger than $1 - n^{-11}$,
\beq
    \label{eqb.9}
    \max \left\{ \sum_{i=1}^{n} \mathbf{A}_{ij} : j \in [n] \right\} \leq \left[ (d-1) \beta_2 + 1 \right] n \theta_n.
\eeq
\indent Due to Lemma \ref{lemb.3} and the uniform upper tail property of $\mathbf{A}$, the similarity matrix $\mathbf{A}$ has the $\textnormal{DP}(\theta_n, \kappa_1, \kappa_2)$ with prob. at least $1 - n^{-11}$, where $\kappa_1 = 4e^2$ and $\kappa_2 = 120d^2$. Note that the absolute constant $\beta_2 > 0$ depends only on $\alpha$ and $d$. Now, let $\mathcal{E}_1$ be the event that $\mathbf{A}$ satisfies the $\textnormal{DP}(\theta_n, \kappa_1, \kappa_2)$ and $\mathcal{E}_2$ denote the event that $\mathbf{A}$ obeys the bound \eqref{eqb.9} on row and column sums. Notice that $\mathbb{P}\{ \mathcal{E}_1 \cap \mathcal{E}_2 \} \geq 1 - 2n^{-11}$. By employing Lemma \ref{lemb.4} on the event $\mathcal{E}_{1} \cap \mathcal{E}_{2}$ with parameters $\lambda = \left[ (d-1) \beta_2 + 1 \right] n \theta_n$, $\delta = \theta_n$, $C = \left[ (d-1) \beta_2 + 1 \right]^{-1}$, $\kappa_1 = 4 e^2$, and $\kappa_2 = 120 d^2$, it proves our claim about (T2):
\beq
    \label{eqb.10}
        \begin{split}
        \textnormal{(T2)} &= \sup_{x \in \mathcal{N}} \left| f_{H(x)}(\mathbf{A}) \right| \leq \sigma \sqrt{\left[(d-1) \beta_2 + 1 \right] n \theta_n},
    \end{split}
\eeq
with probability higher than $1 - 2n^{-11}$, where $\sigma$ is specified as $\sigma = 7680d^2 \left( 1 + \frac{1}{4e^2(1 + \log 2)} \right) + \frac{32}{(d-1) \beta_{2} + 1}(1+4e^2) + 16$. \\
\indent By applying the union bound together with \eqref{eqb.4} and \eqref{eqb.10}, with probability at least $1 - 4n^{-11}$,
\begin{equation*}
    \begin{split}
        &\sup_{x \in \mathcal{N}} \left| x^{\top} \left( \mathbf{A} - \mathbb{E}[\mathbf{A}] \right) x \right| \\
        &\leq \left[ 1 + \beta_1 + \sigma \sqrt{\left( (d-1) \beta_{2} + 1 \right)} \right] \sqrt{n \theta_n}.
    \end{split}
\end{equation*}
\noindent Finally, the bound \eqref{eqb.2} yields with prob. exceeding $1 - 4n^{-11}$,
\begin{equation*}
    \| \mathbf{A} - \mathbb{E}[\mathbf{A}] \| \leq \underbrace{2 \left[ 1 + \beta_1 + \sigma \sqrt{\left( (d-1) \beta_{2} + 1 \right)} \right]}_{c_5} \sqrt{n \theta_n}
\end{equation*}
and note that the universal constant $c_5$ depends only on $\alpha$.

\section{Proof of Technical Lemmas}
\label{appendix:c}

\subsection{Proof of Lemma \ref{lem7.1}}
\label{appendix:c.1}

\indent First, we have by the triangle inequality that
\beq
    \begin{split}
    \label{eqc.1}
    & \ \left\| \mathcal{P}_{T} \left( \mathbf{A} - \mathbb{E}[\mathbf{A}] \right) \right\|_{\infty} \\
    \leq& \ 3 \left( \left\| \mathbf{U} \mathbf{U}^{\top} \left( \mathbf{A} - \mathbb{E}[\mathbf{A}] \right) \right\|_{\infty} \vee \left\| \left( \mathbf{A} - \mathbb{E}[\mathbf{A}] \right) \mathbf{U} \mathbf{U}^{\top} \right\|_{\infty} \right).
    \end{split}
\eeq
Owing to $\left\| \mathbf{U} \mathbf{U}^{\top} \left( \mathbf{A} - \mathbb{E}[\mathbf{A}] \right) \right\|_{\infty} = \left\| \left( \mathbf{A} - \mathbb{E}[\mathbf{A}] \right) \mathbf{U} \mathbf{U}^{\top} \right\|_{\infty}$, it suffices to derive a bound on $\left\| \mathbf{U} \mathbf{U}^{\top} \left( \mathbf{A} - \mathbb{E}[\mathbf{A}] \right) \right\|_{\infty}$. \\
\indent Suppose that the node $i \in [n]$ belongs to the $a^{\textnormal{th}}$ community, \textit{i.e.}, $\Phi^*(i) = a$. Then, for each $j \in [n]$, the $(i, j)^{\textnormal{th}}$ entry of $\mathbf{U} \mathbf{U}^{\top} \left( \mathbf{A} - \mathbb{E}[\mathbf{A}] \right)$ can be expressed as
\begin{equation*}
    \begin{split}
        & \ \left[ \mathbf{U} \mathbf{U}^{\top} \left( \mathbf{A} - \mathbb{E}[\mathbf{A}] \right) \right]_{ij} \\
        =& \ \frac{1}{s_a} \sum_{l \in C_{a}^* \setminus \{ j \}} \left( \mathbf{A}_{lj} - \mathbb{E}[\mathbf{A}_{lj}] \right) \\
        =& \ \frac{1}{s_a} \sum_{l \in C_{a}^* \setminus \{ j \}} \left[ \sum_{e \in \mathcal{E} : \{ l, j \} \subseteq e} \left( W_e - \mathbb{E}[W_e] \right) \right] \\
        =& \ \frac{1}{s_a} \sum_{e \in \mathcal{E}_{j}^{a}} \underbrace{\left| \left( e \cap C_{a}^* \right) \setminus \{ j \} \right| \cdot \left( W_e - \mathbb{E}[W_e] \right)}_{V_e},
    \end{split}
\end{equation*}
where $\mathcal{E}_{j}^{a} := \left\{ e \in \mathcal{E} : j \in e,\ \left( e \cap C_{a}^* \right) \setminus \{ j \} \neq \varnothing \right\}$ for $a \in [k]$ and $j \in [n]$. The random variables $V_e$, $e \in \mathcal{E}_{j}^{a}$, are independent and mean-zero. We can make some remarkable observations:
\begin{itemize}
    \item $\left| V_e \right| \leq d-1$ for all $e \in \mathcal{E}_{j}^{a}$.
    \item The sum of second moments of $V_e$'s is bounded by:
    \beq
        \label{eqc.2}
        \begin{split}
            \sum_{e \in \mathcal{E}_{j}^{a}} \mathbb{E}[V_{e}^2] &\overset{\textnormal{(a)}}{\leq} (d-1)^2 \sum_{e \in \mathcal{E}_{j}^{a}} \textnormal{Var}[W_e]\\
            &\overset{\textnormal{(b)}}{\leq} (d-1)^2 \sum_{e \in \mathcal{E}_{j}^{a}} \mathbb{E}[W_e] \\ &\overset{\textnormal{(c)}}{\leq} (d-1)^2 \sum_{l \in C_{a}^* \setminus \{j\}} \left( \sum_{e \in \mathcal{E} : \left\{ l, j \right\} \subseteq e} \mathbb{E}[W_e] \right)\\
           & \leq (d-1)^2 s_a \underbrace{\binom{n-2}{d-2} p_n}_{\nu_n}
        \end{split}
    \eeq
    where the step (a) attributes to the property $V_{e}^2 \leq (d-1)^2 \left( W_e - \mathbb{E}[W_e] \right)^2$ for every $e \in \mathcal{E}_{j}^{a}$, (b) is due to the property that $W_e \in [0, 1]$, $\forall e \in \mathcal{E}$, and the step (c) is obtained from the set relation
    \begin{equation*}
        \begin{split}
        \mathcal{E}_{j}^{a} &= \left\{ e \in \mathcal{E} : j \in e,\ e \cap \left( C_{a}^* \setminus \{ j \} \right) \neq \varnothing \right\} \\
        &= \bigcup_{l \in C_{a}^* \setminus \{ j \}} \left\{ e \in \mathcal{E} : \left\{ l, j \right\} \subseteq e \right\}.
        \end{split}
    \end{equation*}
\end{itemize}
\indent The two-sided Bernstein's inequality gives with probability at least $1 - 2n^{-13}$,
\begin{equation*}
    \begin{split}
        & \ s_a \left| \left[ \mathbf{U} \mathbf{U}^{\top} \left( \mathbf{A} - \mathbb{E}[\mathbf{A}] \right) \right]_{ij} \right| \\
        \overset{\textnormal{(d)}}{\leq}& \ 9(d-1) \log n + (d-1) \sqrt{26 s_a \nu_n \log n},
    \end{split}
\end{equation*}
where the step (d) is due to the bound \eqref{eqc.2}. Thus, it follows that for every $i, j \in [n]$,
\begin{equation}
    \label{eqc.3}
    \begin{split}
        \left| \left[ \mathbf{U} \mathbf{U}^{\top} \left( \mathbf{A} - \mathbb{E}[\mathbf{A}] \right) \right]_{ij} \right|
        \leq (d-1) \left( \frac{9 \log n}{s_{\min}} + \sqrt{\frac{26 \nu_n \log n}{s_{\min}}} \right)
    \end{split}
\end{equation}
with probability exceeding $1 - 2n^{-13}$. It's straightforward from the condition \eqref{eq14} that
\beq
    \label{eqc.4}
    \left( \log n \right)^2 \leq \frac{1}{c_1} \nu_n s_{\min} \log n.
\eeq
Employing the inequality \eqref{eqc.4} to \eqref{eqc.3}, one has with probability greater than $1 - 2n^{-13}$,
\begin{equation*}
    \begin{split}
        & \ \left| \left[ \mathbf{U} \mathbf{U}^{\top} \left( \mathbf{A} - \mathbb{E}[\mathbf{A}] \right) \right]_{ij} \right| \\
        \leq& \ (d-1) \left( \frac{9}{\sqrt{c_1}} + \sqrt{26} \right) \sqrt{\frac{\nu_n \log n}{s_{\min}}} \\
        \overset{\textnormal{(e)}}{\leq}& \ (d-1) \left( \frac{9}{c_1} + \sqrt{\frac{26}{c_1}} \right) \binom{s_{\min}-2}{d-2} \left( p_n - q_n \right),
    \end{split}
\end{equation*}
where (e) is due to the condition \eqref{eq14}. From the union bound, with probability at least $1 - 2n^{-11}$,
\beq
    \label{eqc.5}
    \begin{split}
    & \ \left\| \mathbf{U} \mathbf{U}^{\top} \left( \mathbf{A} - \mathbb{E}[\mathbf{A}] \right) \right\|_{\infty}\\
    \leq& \ (d-1) \left( \frac{9}{c_1} + \sqrt{\frac{26}{c_1}} \right) \binom{s_{\min}-2}{d-2} \left( p_n - q_n \right).
    \end{split}
\eeq
Finally, putting \eqref{eqc.1} and \eqref{eqc.5} together completes the proof.

\subsection{Proof of Lemma \ref{lemb.1}}
\label{appendix:c.2}

\indent The spectral norm of an $n \times n$ real symmetric matrix can be written as $\left\| \mathbf{M} \right\| = \sup_{x \in \mathbb{S}^{n-1}} \left| \langle \mathbf{M}x, x \rangle \right|$. So, the lower bound part of the inequality \eqref{eqb.1} is obvious. Now, it remains to show the upper bound part of \eqref{eqb.1}. We fix any point $x \in \mathbb{S}^{n-1}$ and take a $x_0 \in \mathcal{N}$ such that $\left\| x - x_0 \right\|_{2} \leq \epsilon$. It follows by the triangle inequality that
\begin{equation*}
    \begin{split}
        & \ \left| \langle \mathbf{M}x, x \rangle \right| - \left| \langle \mathbf{M}x_0, x_0 \rangle \right| \\
        \leq& \ \left| \langle \mathbf{M}x, x \rangle - \langle \mathbf{M}x_0, x_0 \rangle \right| \\
        =& \ \left| \langle \mathbf{M}x, x - x_0 \rangle + \langle \mathbf{M}(x - x_0), x_0 \rangle \right| \\
        \leq& \ \left\| \mathbf{M} \right\| \cdot \|x\|_{2} \cdot \left\| x - x_0 \right\|_{2} + \left\| \mathbf{M} \right\| \cdot \left\| x - x_0 \right\|_{2} \cdot \left\| x_0 \right\|_{2} \\
        \leq& \ 2 \varepsilon \left\| \mathbf{M} \right\|.
    \end{split}
\end{equation*}
Thus, we obtain that for any $x \in \mathbb{S}^{n-1}$,
\beq
    \label{eqc.6}
    \left| \langle \mathbf{M}x, x \rangle \right| - 2 \varepsilon \left\| \mathbf{M} \right\| \leq \sup \left\{ \left| \langle \mathbf{M}y, y \rangle \right| : y \in \mathcal{N} \right\}.
\eeq
By taking supremum to left-hand side of \eqref{eqc.6} over $x \in \mathbb{S}^{n-1}$, we may conclude that
\beq
    \label{eqc.7}
    \left( 1 - 2 \varepsilon \right) \left\| \mathbf{M} \right\| \leq \sup \left\{ \left| \langle \mathbf{M}y, y \rangle \right| : y \in \mathcal{N} \right\}.
\eeq
Dividing $1 - 2 \varepsilon$ from both sides of \eqref{eqc.7}, we get the upper bound on the spectral norm $\left\| \mathbf{M} \right\|$ of an $n \times n$ real symmetric matrix $\mathbf{M}$. This completes the proof of Lemma \ref{lemb.1}.

\bibliographystyle{plain}

\begin{IEEEbiographynophoto}{Jeonghwan Lee}
is currently an undergraduate student studying at the Department of Mathematical Sciences, Korea Advanced Institute of Science and Technology (KAIST). He worked in the Republic of Korea Air Force for his military service from October 2018 to August 2020 as an aerographer. His research interests include information theory, high-dimensional statistics and statistical inference in networks.

\end{IEEEbiographynophoto}

\begin{IEEEbiographynophoto}{Daesung Kim}
received the B.S. degree in electrical engineering from the Korea Advanced Institute of Science and Technology (KAIST), Daejeon, Korea, in 2017. He is currently working toward unified Master and Ph.D. degree in the Department of Electrical Engineering, KAIST. His research interests include information theory, high-dimensional statistics and machine learning.
\end{IEEEbiographynophoto}

\begin{IEEEbiographynophoto}{Hye Won Chung}
(S'08--M'15) received the B.S. degree (with summa cum laude) from Korea Advanced Institute of Science and Technology (KAIST) in 2007, and the M.S. and Ph.D. degrees from Massachusetts Institute of Technology (MIT) in 2009 and 2014, respectively, all in Electrical Engineering and Computer Science. From August 2014 to May 2017, she worked as a Research Fellow in the Department of Electrical Engineering and Computer Science at the University of Michigan. Since June 2017, she has been working as an assistant professor at KAIST. Her research interests include information theory, statistical inference, machine learning and quantum optical communications. 
\end{IEEEbiographynophoto}

\end{document}